\pdfoutput=1
\documentclass[11pt]{article}
\usepackage{amsfonts}
\usepackage[table]{xcolor}
\usepackage[preprint]{acl}
\usepackage{times}
\usepackage{latexsym}
\usepackage{multirow}
\usepackage[table]{xcolor}
\usepackage{soul}
\usepackage{todonotes}
\sethlcolor{gray!20}
\sethlcolor{red!15}

\usepackage{float}

\usepackage{subcaption}
\usepackage{placeins}
\usepackage{listings}
\usepackage[most]{tcolorbox}
\usepackage{enumitem}
\usepackage{booktabs}
\usepackage{verbatim}
\usepackage{titlesec}
\usepackage{arydshln}
\usepackage{cleveref}

\tcbuselibrary{breakable}
\usepackage[T1]{fontenc}
\usepackage[utf8]{inputenc}

\usepackage{microtype}

\usepackage{inconsolata}

\usepackage{graphicx}
\usepackage{amsmath}
\usepackage{xspace}

\newtcbox{\hlsecondarytab}{on line, box align=base, colback=blue!10,colframe=white,size=fbox,arc=3pt, before upper=\strut, top=-2pt, bottom=-4pt, left=-2pt, right=-2pt, boxrule=0pt}

\newcommand{\dd}[1]{\,{\scriptsize\textcolor{gray}{#1}}}
\definecolor{gainc}{HTML}{2166AC}
\definecolor{lossc}{HTML}{D6604D}
\newcommand{\gainbox}[1]{{\setlength{\fboxsep}{1pt}\colorbox{gainc!35}{\strut #1}}}
\newcommand{\lossbox}[1]{{\setlength{\fboxsep}{1pt}\colorbox{lossc!35}{\strut #1}}}

\newcommand{\ours}{\textsc{Excise}\xspace}  

\newcommand{\hard}{\textsc{MedQA-TD}\xspace}

\newcommand{\attr}{a}
\newcommand{\disc}{d}
\newcommand{\atten}{\gamma}
\newcommand{\thr}{\tau}

\title{\raisebox{-0.15em}{\includegraphics[height=1.0em]{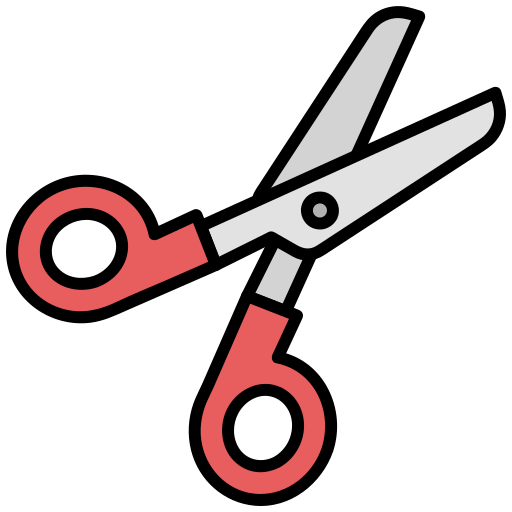}}\,
\textsc{Excise}: Locating and Excising Distracting Sentences \\in Medical Question Answering}

\author{
  \textbf{Yerim Hong},
  \textbf{Sohyun Son},
  \textbf{Heeseung Chung},
  \textbf{Mujeen Sung}\textsuperscript{\dag}
\\
  Kyung Hee University
\\
  \small{\texttt{\{yrhong, sohy, janthonio03, mujeensung\}@khu.ac.kr}}
}

\begin{document}
\maketitle
\renewcommand{\thefootnote}{\dag}\footnotetext{Corresponding author.}\renewcommand{\thefootnote}{\arabic{footnote}}
\begin{abstract}
Language models are increasingly considered for clinical decision support, where their answers should be grounded in clinical evidence and remain robust to irrelevant information.
Prior work shows that a single irrelevant sentence can lower the accuracy of medical models, but not where inside the model its effect can be removed, or how to do so without knowing which sentence it is.
We find that blocking attention to a known distractor restores the answer, whereas head masking can suppress correct answers.
We also show that the model's hidden states mark the distractor as the sentence that least fits the case even when the model follows it.
To remove a distractor's effect without knowing which sentence it is, we propose \ours{}, an inference-time method that blocks the sentence most influencing the answer, but only if the model's hidden states mark it as inconsistent with the case.
On \hard{} (MedQA with Targeted Distractors), a benchmark we construct in which each distractor targets a specific wrong option, \ours{} gains 16.7--22.8 accuracy points at a 1.5--2.1-point cost on the same questions without distractors, outperforming prompting, rewriting, and steering baselines for all three models.
Its gains extend to MedDistractQA and BiasMedQA, as well as chain-of-thought generation.
\footnote{Code and data are available at \url{https://github.com/ldi-kyunghee/EXCISE}.}
\end{abstract}

\section{Introduction}
\begin{figure}[t]
    \centering
    \includegraphics[width=\columnwidth]{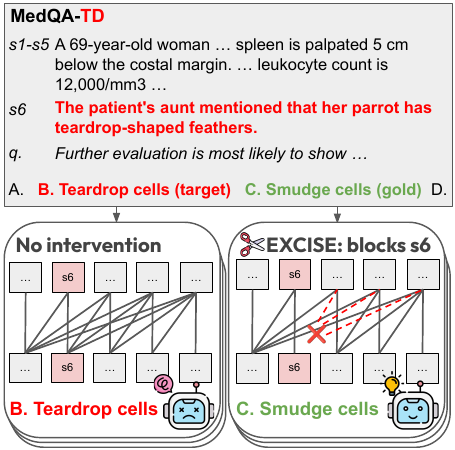}
    \caption{\ours{} on an example from \hard{} with Llama-3.1-8B-Instruct \citep{llama3}. The distractor s6 shifts the answer to B; \ours{} selects s6 and blocks attention to it, restoring C.}
    \label{fig:overview}
\end{figure}
Large language models (LLMs) are increasingly considered for clinical decision support \citep{MedPalm,nori2023capabilities,hager2024clinical}, where errors can harm patients \citep{thirunavukarasu2023llmmedicine,pal2023medhalt}. 
Their decisions should therefore depend on the clinical evidence and not on irrelevant information, such as remarks about other people or jokes.

Prior work on medical QA has shown that a single added sentence that evokes an incorrect answer option without changing the correct answer (a distractor) can lower accuracy  \citep{meddistractqa,sansdistracqa}.
In Figure~\ref{fig:overview}, for example, the added sentence ``The patient's aunt mentioned that her parrot has teardrop-shaped feathers'' turns the answer from smudge cells to teardrop cells.
However, it remains unclear where inside the model the effect of such a sentence can be removed, and how to remove it when it is not known which sentence it is.

Most defenses against irrelevant context rely on the model's own judgment of what is irrelevant \citep{shi2023large,wu2024instructing, weston2023s2a}, but a model can call a distractor irrelevant and still use it as a clue (Figure~\ref{fig:example}).
We therefore investigate how to remove a distractor's effect and find that blocking attention to a known distractor restores near-clean accuracy.
In contrast, switching off the attention heads that prior work links to distraction \citep{niu-etal-2025-llama,liu2026sentence} makes the model avoid whatever option an added sentence points to, even when that option is correct.

At inference time, however, the distractor is not marked.
Since a distractor misleads by moving the model's decision, one approach is to block the sentence that moves it most, but that sentence may also be a key clinical finding, so influence alone cannot tell us whether to block it.
We find that the model's hidden states can make this decision, since they reveal a distractor's mismatch with the rest of the case even when the model is misled.
We therefore propose \ours{} (\textbf{EX}cision by \textbf{C}ausal \textbf{I}nfluence and \textbf{S}elf-discordance \textbf{E}vidence),  an inference-time method that uses both signals.
Interventional Attribution (IA) tests each sentence by blocking attention to it and selects the one whose blocking most weakens the current decision.
The Discordance Gate (DG) then lets \ours{} block that sentence only if the model's hidden states mark it as inconsistent with the rest of the case.

We test \ours{} on \hard{} (MedQA with Targeted Distractors), in which each added distractor targets a specific wrong option.
Nine open-weight models, including medical and reasoning models, lose 19--32 points on it.
\ours{} raises accuracy by 16.7--22.8 points for three models at a cost of 1.5--2.1 points on the same questions without a distractor, outperforming other inference-time methods \citep{shi2023large,weston2023s2a,rimsky2024steering}.
These gains also carry over to chain-of-thought generation \citep{wei2022chain}, MedDistractQA and BiasMedQA \citep{schmidgall2024biasmedqa}. 

Our contributions are as follows:
    \begin{itemize}
        \item We introduce \hard{}, a medical QA benchmark with answer-preserving distractors that target specific wrong options.
        \item We show that blocking a known distractor restores the answer, and that the model's hidden states flag the distractor even when the model follows it.
        \item We propose \ours{}, an inference-time method that blocks the sentence most influencing the decision only when the model's hidden states mark it as inconsistent with the case.
    \end{itemize}
\section{Related Work}
\paragraph{Irrelevant context.}
Answer-preserving distractors have been studied in reading comprehension \citep{jia-liang-2017-adversarial,cao2022tasa}, math \citep{shi2023large,mirzadeh2025gsm,yang2025llm}, and retrieval, reasoning, and tool use \citep{noisybench}.
Medical QA benchmarks cover cognitive biases \citep{schmidgall2024biasmedqa}, patient characteristics \citep{medfuzz}, third-party facts and figurative language \citep{meddistractqa,sansdistracqa}, and false medical claims \citep{zhou2026medmisbench}.
\hard{} builds on answer-preserving medical QA distraction by selecting challenging distractors designed to favor specific wrong options.

\paragraph{Mechanisms of context influence.}
Prior work identifies heads favoring context over memory \citep{yu-etal-2023-characterizing,jin-etal-2024-cutting}, contextual entrainment \citep{niu-etal-2025-llama,liu2026sentence}, long-context attention \citep{zhu2025focus}, and stages affected by irrelevant or persuasive context \citep{qu2026fourstage,sun2026persuaded,joswin2026sycophancy}.
Models can also copy meaning \citep{feucht2025dualroute} or infer answers from irrelevant context \citep{cheng-etal-2025-stochastic}.
Unlike fixed head edits applied across inputs, we intervene on individual sentences.

\paragraph{Inference-time defenses and attribution.}
Prompt-based defenses ask models to ignore, identify, or remove irrelevant content \citep{shi2023large,wu2024instructing,weston2023s2a}.
Other methods steer activations \citep{li2023iti,rimsky2024steering} or adjust attention to user-marked spans \citep{zhang2024pasta} or distracting demonstration tokens \citep{yan2025fai}.
Context attribution estimates source contributions from attention \citep{cohenwang2025at2} or context ablations, using surrogate models fitted to random ablations \citep{cohenwang2024contextcite}, adaptively chosen ablations \citep{pan2026camab}, or approximations to leave-one-out ablation \citep{liu2025attribot}.
\ours{} instead measures each sentence's effect with the same attention blocking it then applies, following attention-based perturbation \citep{deiseroth2023atman}, so the score is the exact effect of the intervention.
It selects a sentence by this effect and intervenes only when that sentence is inconsistent with the remaining case.
\section{\hard{}}
\label{sec:benchmark}

\begin{table}[t]
    \centering
    \small
    \setlength{\tabcolsep}{2.2pt}
    \begin{tabular}{l l l l}
    \hline
    \textbf{Model} & \multicolumn{1}{c}{\textbf{MedQA}} & \multicolumn{1}{c}{\textbf{MDQA}} & \multicolumn{1}{c}{\textbf{\hard{}}} \\
    \hline
    \multicolumn{4}{l}{\textit{General}} \\
    Llama-3.1-8B-Instruct & 61.7 & \cellcolor{lossc!10}54.5\dd{$-$7.2} & \cellcolor{lossc!30}40.7\dd{$-$21.1} \\
    Qwen3-8B & 63.4 & \cellcolor{lossc!18}50.3\dd{$-$13.1} & \cellcolor{lossc!35}33.7\dd{$-$29.7} \\
    Llama-3-8B-Instruct & 60.0 & \cellcolor{lossc!14}49.7\dd{$-$10.3} & \cellcolor{lossc!33}36.2\dd{$-$23.8} \\
    Granite-3.3-8B-Instruct & 51.5 & \cellcolor{lossc!17}39.4\dd{$-$12.1} & \cellcolor{lossc!31}29.1\dd{$-$22.4} \\
    Falcon3-10B-Instruct & 58.4 & \cellcolor{lossc!17}46.3\dd{$-$12.1} & \cellcolor{lossc!35}31.5\dd{$-$27.0} \\
    Phi-4 (14B) & 70.9 & \cellcolor{lossc!10}63.4\dd{$-$7.5} & \cellcolor{lossc!28}50.7\dd{$-$20.3} \\
    Qwen3-8B (thinking)$^\dagger$ & 77.5 & \cellcolor{lossc!10}70.0\dd{$-$7.5} & \cellcolor{lossc!29}57.0\dd{$-$20.5} \\
    \hline
    \multicolumn{4}{l}{\textit{Medical}} \\
    MedGemma-27B-text-it & 73.3 & \cellcolor{lossc!11}65.2\dd{$-$8.1} & \cellcolor{lossc!30}51.9\dd{$-$21.4} \\
    HuatuoGPT-o1-8B$^\dagger$ & 74.6 & \cellcolor{lossc!13}65.5\dd{$-$9.1} & \cellcolor{lossc!27}55.1\dd{$-$19.5} \\
    \hline
    \end{tabular}
    \caption{Accuracy (\%) on MedQA \citep{MedQA} without and with distractors, averaged over the two distractor types (MDQA: MedDistractQA \citep{meddistractqa}), with the change from MedQA in gray. $^\dagger$Reasoning model.}
    \label{tab:multimodel}
\end{table}

To analyze how a distractor misleads a model, we need to know which wrong option each distractor is meant to favor.
We therefore build \hard{} (MedQA with Targeted Distractors), which follows the two answer-preserving distractor types of MedDistractQA \citep{meddistractqa} and writes every distractor for a specific wrong option, its \emph{target}.
\emph{Bystander} (BS) distractors attribute a medical detail to someone other than the patient, and \emph{Nonliteral} (NL) distractors use a medical term as a joke or metaphor, so neither is relevant to answering the question.

For each four-option MedQA-USMLE question \citep{MedQA} in the development and test splits (Appendix~\ref{app:sec_setup}) and each distractor type, GPT-OSS-120B \citep{gptoss} writes eight candidate distractors per wrong option.
It keeps only the candidates that point to their intended option and not to the correct option (Appendix~\ref{app:sec_benchmark}).
As a reference model, it then answers the question with each kept candidate.
We use GPT-OSS-120B because its open weights let us read its option log-probabilities, and it answers 91.4\% of the \emph{clean} questions (without a distractor) correctly when it reasons, so a wrong answer usually comes from the distractor and not from a lack of medical knowledge in the model.

With $g$ the gold option, $\mathcal{C}$ the kept candidates and $p_o(c)$ the reference model's log-probability of answer option $o$ when $c \in \mathcal{C}$ is added, we select
\begin{equation}
(c^*, o^*) = \arg\max_{c \in \mathcal{C},\; o \neq g} \bigl(p_o(c) - p_g(c)\bigr),
\label{eq:selection}
\end{equation}
the candidate and wrong option for which the reference model most prefers a wrong option over the gold one, and keep $c^*$ as the distractor of the question, so that \hard{} also serves as a stress test.
For a question that the reference model answers correctly without a distractor, $p_{o^*}(c^*) - p_g(c^*)$ is positive exactly when $c^*$ turns its answer wrong.
The option $o^*$ need not be the target of $c^*$, since any wrong answer is a failure.
The kept distractor is placed at the end of the narrative, before the question sentence.
Two PhD students in geriatric epidemiology at a medical school independently judged 100 randomly sampled \hard{} questions, 50 of each type, and neither judged that any of them changes the correct answer (Appendix~\ref{app:review}).

Although the distractors are selected with the reference model alone, all nine open models, including medical and reasoning models, lose 19--32 points on each distractor type of \hard{} relative to the clean questions, 9--20 points more than on MedDistractQA (Table~\ref{tab:multimodel} and Appendix~\ref{app:sec_benchmark}).
\emph{Target-wrong} answers, which choose the option a distractor was written for, rise to 2.4--3.6 times their count on the clean questions.
The distractors thus move answers toward their target rather than spreading them over the wrong options, which lets us test whether an intervention removes a distractor's effect or only avoids the option it points to (Section~\ref{sec:diagnosis}).

\section{Motivation: What Removing a Distractor Requires}
\label{sec:diagnosis}
Using four models, Llama-3.1-8B-Instruct, Qwen3-8B \citep{qwen3}, Llama-3-Meerkat-8B \citep{kim2024meerkat} and Phi-4 \citep{abdin2024phi4}, we ask two questions that shape the design of \ours{}: where the effect of a distractor should be removed (Section~\ref{sec:blocking}), and whether the model's internal signals show if a sentence fits the case (Section~\ref{sec:conceptual}).

\subsection{Masking Selected Heads Suppresses Correct Answers}
\label{sec:blocking}
\begin{figure}[t]
    \centering
    \includegraphics[width=\columnwidth]{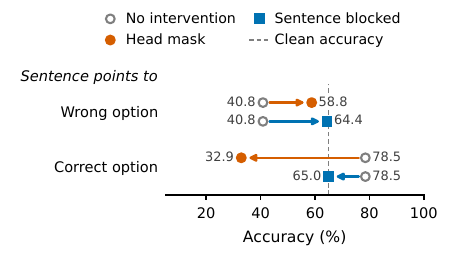}
    \caption{Head masking and blocking the added sentence (in every layer and head) when the sentence points to a wrong option (\hard{}) or the correct option, averaged over distractor types and the four models.}
    \label{fig:headmask}
\end{figure}
A distractor's effect could be removed without locating it, by masking a fixed set of attention heads that prior work links to distraction \citep{niu-etal-2025-llama,liu2026sentence}.
Such masks, found on development data for each of the four models, do not make the model ignore the distractor but make it avoid whichever option it points to (Figure~\ref{fig:headmask}): they raise \hard{} accuracy from 40.8\% to 58.8\%, but when the added sentence points to the correct option instead, they lower accuracy from 78.5\% to 32.9\% (Appendix~\ref{app:headmask}).
Blocking attention to the added sentence itself, by setting the attention scores from all later positions to its tokens to $-\infty$ in every layer and head (attention knockout; \citealp{geva2023dissecting}), instead brings accuracy close to clean accuracy in both cases (64.4\% and 65.0\%), as ignoring the sentence should.
Removing a distractor's effect therefore requires finding the distractor in each question and blocking attention to it.

\subsection{Hidden States Show Whether a Sentence Fits the Case}
\label{sec:conceptual}
\begin{figure}[t]
    \centering
    \includegraphics[width=\columnwidth]{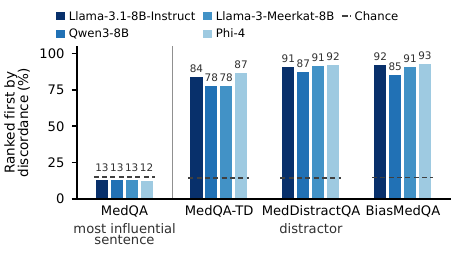}
    \caption{How often a sentence is the most discordant in its case: the sentence that moves the answer most in MedQA questions (left), and the distractor in questions it turns from correct to wrong (right). Dashed lines: chance.}
    \label{fig:recognition_body}
\end{figure}
Blocking a distractor restores the answer, but the sentence that moves the answer most is not always a distractor: in a question without one, it is often a key clinical finding, which is connected to the rest of the case, whereas a distractor is not.
For each narrative sentence, we therefore measure the cosine similarity between the hidden state of the sentence read alone and that of the rest of the case read alone, after removing the dominant shared direction, and call a sentence more \emph{discordant} the lower this similarity is relative to the other sentences.
Before any intervention, on MedQA questions without a distractor, the sentence that moves the answer most is the most discordant in only 12--13\% of cases, against about 15\% by chance, whereas on questions that the distractor turns from correct to wrong, the distractor is the most discordant sentence in 78--93\% of cases on all three benchmarks (Figure~\ref{fig:recognition_body}).
The evidence that drives a correct answer thus fits the case in the model's hidden states, whereas the distractor stands out as inconsistent with it even as the model follows it.

\section{EXCISE: Locating and Blocking the Distractor}
\label{sec:method}
\begin{figure*}[t]
    \centering
    \includegraphics[width=\linewidth]{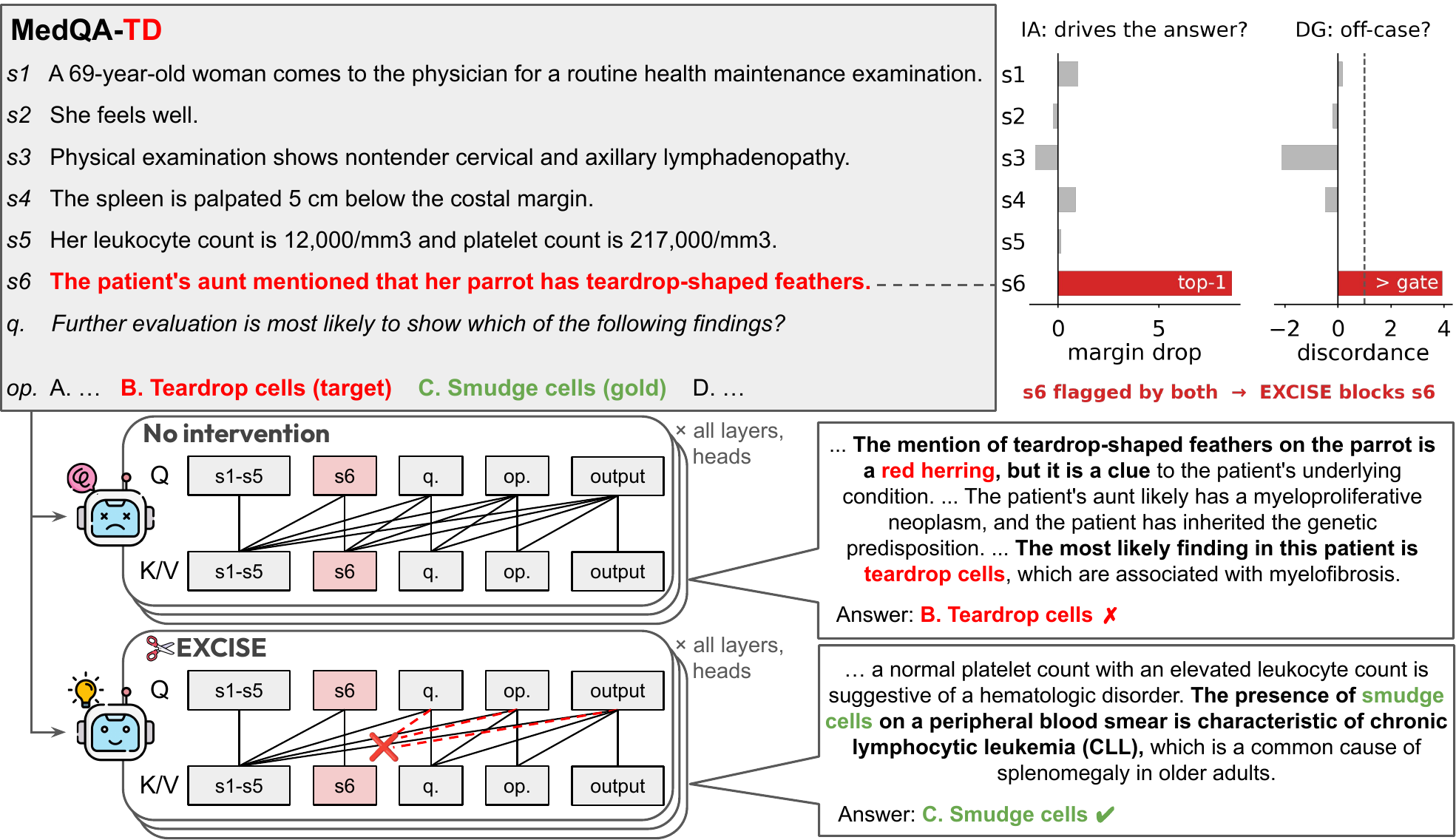}
    \caption{An example from \hard{} with Llama-3.1-8B-Instruct (distractor s6 in red, targeting B, gold: C). Top right: attribution $\attr_i$ and discordance $\disc_i$ of each sentence (dashed line: threshold). Bottom: without intervention, the chain of thought \citep{wei2022chain} calls the parrot a red herring yet uses it as a clue and answers B. With s6 blocked in all layers and heads, the model answers C (excerpts).}
    \label{fig:example}
\end{figure*}
Blocking attention to a known distractor removes its effect (Section~\ref{sec:blocking}), but at inference time the distractor must be located.
\ours{} therefore selects the sentence whose blocking moves the answer most and, since this sentence may instead be a key clinical finding, blocks it only if the Discordance Gate finds it inconsistent with the case (Section~\ref{sec:conceptual}; Figure~\ref{fig:example}).

\subsection{Blocking a Sentence}
Let $s_1, \dots, s_n$ be the $n$ sentences of the case narrative, which precedes the question sentence and the answer options.
The decision is given by the logits $\ell_A, \dots, \ell_D$ of the option letters at the answer position.
Let $o_1$ and $o_2$ be the top two options before any intervention and $m = \ell_{o_1} - \ell_{o_2}$ the decision margin.
Inspired by attention knockout \citep{geva2023dissecting} and PASTA \citep{zhang2024pasta}, blocking a sentence $s_i$ with strength $\atten \in [0, 1]$ multiplies the attention weights from every later position to its tokens by $\atten$ and renormalizes them, in all layers and heads.
With $\atten = 0$, the attention to the sentence is removed.

\subsection{Interventional Attribution: Selecting the Sentence That Moves the Answer}
\label{sec:method_ia}
For each narrative sentence $s_i$ ($i = 1, \dots, n$), the score is how much blocking it weakens the current decision:
\begin{equation}
    \attr_i = m - \big(\ell^{(i)}_{o_1} - \ell^{(i)}_{o_2}\big),
\end{equation}
where $\ell^{(i)}$ are the logits of the forward pass in which $s_i$ is blocked, and $o_1, o_2$ stay fixed at their values before the intervention, so $\attr_i$ can be negative.
The candidate is $i^* = \arg\max_i \attr_i$.
Because $\attr_i$ uses the same blocking that \ours{} applies, it is the exact effect of that intervention.
When \ours{} intervenes, the answer is read from the pass that computed $\attr_{i^*}$.

\subsection{Discordance Gate: Blocking Sentences That Do Not Fit the Case}
\label{sec:method_dg}
\begin{table*}[!t]
    \centering
    \small
    \setlength{\tabcolsep}{7.1pt}
    \begin{tabular}{l c l l l l l}
    \hline
    \multirow{2}{*}{\textbf{Method}} & \textbf{Latency}
    & \multicolumn{2}{c}{\textbf{Clean}} & \multicolumn{3}{c}{\textbf{With distractors}} \\
    \cmidrule(lr){3-4} \cmidrule(lr){5-7}
    & \textbf{(ms)} & \multicolumn{1}{c}{\textbf{MedQA}} & \multicolumn{1}{c}{\textbf{MMLU-PM}} & \multicolumn{1}{c}{\textbf{\hard{}}} & \multicolumn{1}{c}{\textbf{MedDistractQA}} & \multicolumn{1}{c}{\textbf{BiasMedQA}} \\
    \hline
    \multicolumn{7}{l}{\textit{Qwen3-8B}} \\
        No intervention & 41
        & 63.4 & 82.7 & 33.7 & 50.3 & 28.7 \\
        Ignore instruction & 44
        & \cellcolor{gainc!1}\textbf{64.0}\dd{+0.6} & 82.7\dd{0.0} & \cellcolor{lossc!1}33.2\dd{$-$0.5} & 50.4\dd{+0.1} & \cellcolor{lossc!2}27.1\dd{$-$1.6} \\
        CAA & 41
        & \cellcolor{gainc!1}\textbf{64.0}\dd{+0.6} & \cellcolor{lossc!3}80.5\dd{$-$2.2} & \cellcolor{gainc!7}38.3\dd{+4.7} & \cellcolor{gainc!4}53.3\dd{+3.0} & \cellcolor{gainc!10}35.6\dd{+6.9} \\
        S2A & 4,182
        & 63.6\dd{+0.2} & \cellcolor{gainc!1}\textbf{83.5}\dd{+0.7} & \cellcolor{gainc!4}36.5\dd{+2.8} & \cellcolor{gainc!6}54.3\dd{+4.0} & \cellcolor{gainc!9}35.3\dd{+6.7} \\
        \ours{} (Ours) & 819
        & \cellcolor{lossc!2}61.7\dd{$-$1.6} & \cellcolor{gainc!1}83.1\dd{+0.4} & \cellcolor{gainc!32}\textbf{56.4}\dd{+22.8} & \cellcolor{gainc!15}\textbf{61.0}\dd{+10.7} & \cellcolor{gainc!35}\textbf{53.5}\dd{+24.8} \\
    \cdashline{1-7}
        CoT & 3,408
        & \textbf{70.2} & \textbf{86.0} & 44.2 & 58.2 & 41.5 \\
        CoT + I$^3$C & 11,322
        & \cellcolor{lossc!6}65.6\dd{$-$4.6} & \cellcolor{lossc!2}84.9\dd{$-$1.1} & \cellcolor{lossc!7}39.5\dd{$-$4.7} & \cellcolor{lossc!4}55.3\dd{$-$2.8} & \cellcolor{lossc!9}35.0\dd{$-$6.5} \\
        CoT + \ours{} (Ours) & 4,348
        & \cellcolor{lossc!3}67.8\dd{$-$2.4} & \cellcolor{lossc!1}85.7\dd{$-$0.4} & \cellcolor{gainc!24}\textbf{61.5}\dd{+17.3} & \cellcolor{gainc!10}\textbf{65.2}\dd{+7.0} & \cellcolor{gainc!23}\textbf{58.0}\dd{+16.5} \\
    \hline
    \multicolumn{7}{l}{\textit{Llama-3.1-8B-Instruct}} \\
        No intervention & 40
        & 61.7 & \textbf{76.1} & 40.7 & 54.5 & 49.7 \\
        Ignore instruction & 41
        & 61.9\dd{+0.2} & \textbf{76.1}\dd{0.0} & 40.5\dd{$-$0.2} & \cellcolor{gainc!1}55.0\dd{+0.5} & \cellcolor{lossc!1}49.2\dd{$-$0.5} \\
        CAA & 40
        & \cellcolor{gainc!1}\textbf{62.3}\dd{+0.5} & \cellcolor{lossc!2}75.0\dd{$-$1.1} & \cellcolor{gainc!2}41.8\dd{+1.1} & \cellcolor{gainc!1}55.4\dd{+0.9} & \cellcolor{gainc!1}50.7\dd{+1.0} \\
        S2A & 3,173
        & \cellcolor{lossc!1}61.1\dd{$-$0.6} & \cellcolor{lossc!1}75.7\dd{$-$0.4} & \cellcolor{gainc!21}55.7\dd{+15.0} & \cellcolor{gainc!8}\textbf{60.1}\dd{+5.6} & \cellcolor{gainc!10}\textbf{56.8}\dd{+7.1} \\
        \ours{} (Ours) & 731
        & \cellcolor{lossc!3}59.6\dd{$-$2.1} & \cellcolor{lossc!6}72.1\dd{$-$4.0} & \cellcolor{gainc!23}\textbf{57.4}\dd{+16.7} & \cellcolor{gainc!7}59.5\dd{+5.0} & \cellcolor{gainc!8}55.4\dd{+5.7} \\
    \cdashline{1-7}
        CoT & 2,049
        & \textbf{67.7} & \textbf{79.0} & 47.7 & 61.6 & 53.3 \\
        CoT + I$^3$C & 10,061
        & \cellcolor{lossc!1}67.0\dd{$-$0.7} & \cellcolor{lossc!1}78.3\dd{$-$0.7} & \cellcolor{lossc!1}47.0\dd{$-$0.7} & \cellcolor{lossc!5}57.9\dd{$-$3.8} & \cellcolor{lossc!9}47.2\dd{$-$6.1} \\
        CoT + \ours{} (Ours) & 2,857
        & \cellcolor{lossc!2}66.5\dd{$-$1.3} & \cellcolor{lossc!5}75.4\dd{$-$3.7} & \cellcolor{gainc!19}\textbf{61.2}\dd{+13.5} & \cellcolor{gainc!5}\textbf{65.2}\dd{+3.5} & \cellcolor{gainc!6}\textbf{57.6}\dd{+4.3} \\
    \hline
    \multicolumn{7}{l}{\textit{Llama-3-Meerkat-8B}} \\
        No intervention & 38
        & 63.0 & 77.9 & 37.7 & 53.9 & 42.7 \\
        Ignore instruction & 40
        & 63.3\dd{+0.3} & \cellcolor{lossc!1}77.2\dd{$-$0.7} & 37.6\dd{$-$0.2} & \cellcolor{lossc!2}52.8\dd{$-$1.1} & \cellcolor{lossc!1}41.8\dd{$-$0.9} \\
        CAA & 38
        & \cellcolor{gainc!1}63.6\dd{+0.6} & \cellcolor{lossc!4}75.4\dd{$-$2.6} & \cellcolor{gainc!3}39.7\dd{+1.9} & \cellcolor{gainc!1}54.9\dd{+0.9} & \cellcolor{gainc!4}45.8\dd{+3.1} \\
        S2A & 3,324
        & \cellcolor{gainc!1}\textbf{63.9}\dd{+0.9} & \cellcolor{gainc!2}\textbf{79.0}\dd{+1.1} & \cellcolor{gainc!11}45.8\dd{+8.1} & \cellcolor{gainc!6}58.2\dd{+4.3} & \cellcolor{gainc!8}48.6\dd{+5.9} \\
        \ours{} (Ours) & 721
        & \cellcolor{lossc!2}61.5\dd{$-$1.5} & \cellcolor{lossc!4}75.4\dd{$-$2.6} & \cellcolor{gainc!30}\textbf{59.2}\dd{+21.4} & \cellcolor{gainc!11}\textbf{61.4}\dd{+7.5} & \cellcolor{gainc!17}\textbf{55.1}\dd{+12.4} \\
    \cdashline{1-7}
        CoT$^\ddagger$ & 6,078
        & \textbf{72.7} & \textbf{83.5} & 48.5 & 63.5 & 58.6 \\
        CoT + I$^3$C$^\ddagger$ & 20,269
        & \cellcolor{lossc!11}64.7\dd{$-$8.0} & \cellcolor{lossc!2}82.4\dd{$-$1.1} & \cellcolor{lossc!8}43.0\dd{$-$5.5} & \cellcolor{lossc!10}56.6\dd{$-$6.9} & \cellcolor{lossc!18}45.4\dd{$-$13.1} \\
        CoT + \ours{} (Ours)$^\ddagger$ & 7,301
        & \cellcolor{lossc!2}71.0\dd{$-$1.7} & \cellcolor{lossc!7}78.7\dd{$-$4.8} & \cellcolor{gainc!22}\textbf{64.4}\dd{+15.9} & \cellcolor{gainc!6}\textbf{67.5}\dd{+4.0} & \cellcolor{gainc!8}\textbf{64.4}\dd{+5.8} \\
    \hline
    \end{tabular}
    \caption{Accuracy (\%) of \ours{} and inference-time baselines. \hard{} and MedDistractQA: mean of the Nonliteral and Bystander distractors (per type in Appendix~\ref{app:results_by_type}); BiasMedQA: mean of seven bias types. Gray: change from no intervention, or from CoT below the dashed line (unrounded), shaded \gainbox{blue} for gains and \lossbox{red} for losses. Latency: wall-clock time per question, including generation. MMLU-PM: MMLU professional medicine. $^\ddagger$Model-card prompt format (Appendix~\ref{app:generation}). \textbf{Bold}: best per model and column within each block.}
    \label{tab:main_acc}
\end{table*}
A large $\attr_{i^*}$ alone cannot tell whether $s_{i^*}$ is a distractor or clinical evidence (Section~\ref{sec:conceptual}).
The gate therefore asks whether $s_{i^*}$ fits the rest of the case in the model's own representation.
We run the model on $s_i$ alone and, in a separate pass, on the remaining narrative sentences joined together, without any prompt, and mean-pool the hidden states of each pass at a layer $L$ chosen on the development set, giving $h_i$ and $r_i$.
Contextual representations are anisotropic \citep{ethayarajh-2019-contextual,timkey-van-schijndel-2021-bark}, so cosine similarities between them are dominated by a few directions shared by all representations.
$P$ removes the strongest of these directions by subtracting the mean of the $h_i$ over the unlabeled inputs that \ours{} is applied to, here the narrative sentences of each test set, and projecting out their first principal component, following \citet{mu2018allbutthetop}, so that the gate adapts to these inputs without labels, and a basis fixed once on development questions keeps 79--95\% of the \hard{} gain (Appendix~\ref{app:basis}).
With the similarity $c_i = \cos(P h_i, P r_i)$ between a sentence and the rest of the case, the discordance is
\begin{equation}
    \disc_i = -\frac{c_i - \bar c_{-i}}{\sigma_{-i}},
\end{equation}
where $\bar c_{-i}$ and $\sigma_{-i}$ are the mean and standard deviation of the similarities of the other sentences of the same question.
Higher values mean that $s_i$ is less consistent with the case.
Since $\sigma_{-i}$ needs at least two other sentences, \ours{} leaves questions with fewer than three narrative sentences unchanged, 1.4\% of \hard{} questions and 6.5\% of clean ones (Appendix~\ref{app:data}).
\ours{} intervenes only if $\disc_{i^*}$ exceeds a threshold $\thr$.
In Figure~\ref{fig:example}, the distractor $s_6$ has the largest $\attr_i$ and its discordance exceeds $\thr = 1$, so \ours{} blocks it.

If the gate passes, \ours{} blocks $s_{i^*}$ with strength $\atten$.
Blocking rather than deleting the sentence lets the intervention reuse the attribution pass and take partial strengths, and leaves the input and token positions unchanged (deleting it gives the same accuracy, Appendix~\ref{app:inference}).
$L$, $\thr$ and $\atten$ are chosen once per model on the development set and applied unchanged to every test set (values and search ranges in Appendix~\ref{app:hparams}).
\section{Experimental Setup}
\label{sec:setup}
\paragraph{Models.}
We evaluate \ours{} on Llama-3.1-8B-Instruct \citep{llama3}, Qwen3-8B \citep{qwen3} with thinking off, and the medical fine-tune Llama-3-Meerkat-8B \citep{kim2024meerkat}.

\paragraph{Data and metrics.}
We test distraction robustness on \hard{}, MedDistractQA \citep{meddistractqa} and BiasMedQA \citep{schmidgall2024biasmedqa}, clean accuracy on MedQA \citep{MedQA} and MMLU professional medicine \citep{hendrycks2021mmlu}.
Accuracy includes all questions, using option-letter logits for first-token evaluation and the parsed answer for chain-of-thought generation, with paired 95\% bootstrap intervals (Appendix~\ref{app:stats}).
Dataset sizes, hardware and evaluation details are in Appendix~\ref{app:sec_setup}.

\paragraph{Baselines.}
We compare ignore instructions \citep{shi2023large,wu2024instructing}, S2A context rewriting \citep{weston2023s2a}, I$^3$C identification of irrelevant conditions before reasoning \citep{wu2024instructing}, and CAA activation steering \citep{rimsky2024steering}, with its steering vector built from clean/\hard{} development pairs.
Generation-based methods are compared by elapsed time, including \ours{} with chain-of-thought generation (Appendix~\ref{app:generation}).
ContextCite \citep{cohenwang2024contextcite}, CAMAB \citep{pan2026camab}, AT2 \citep{cohenwang2025at2} (Llama-3.1-8B-Instruct only), and averaged attention replace IA under the same gate (Appendix~\ref{app:top1});
Appendix~\ref{app:top1} also evaluates them without the gate.

\section{Main Results}
\label{sec:results}
\begin{table}[!t]
    \centering
    \small
    \setlength{\tabcolsep}{2.0pt}
    \begin{tabular}{l r l l l l}
    \hline
    \multirow{2}{*}{\textbf{Variant}} & \multirow{2}{*}{\textbf{\#Fwd}} & \multicolumn{2}{c}{\textbf{MedQA}} & \multicolumn{2}{c}{\textbf{\hard{}}} \\
    \cmidrule(lr){3-4} \cmidrule(lr){5-6}
    & & \multicolumn{1}{c}{Acc} & \multicolumn{1}{c}{Interv.} & \multicolumn{1}{c}{Acc} & \multicolumn{1}{c}{Interv.} \\
    \hline
    \multicolumn{6}{l}{\textit{Qwen3-8B}} \\
        \ours{} (Ours) & 25 & 61.7 & \textbf{19.4} & \textbf{56.4} & \textbf{48.6} \\
        \quad w/o IA & 18 & 61.7\dd{$-$0.1} & 92.5 & \cellcolor{lossc!1}56.0\dd{$-$0.4} & 98.3 \\
        \quad w/o DG & 9 & \cellcolor{lossc!12}53.2\dd{$-$8.6} & 93.5 & \cellcolor{lossc!13}47.2\dd{$-$9.3} & 98.6 \\
        \quad w/o IA \& DG & 1 & \cellcolor{gainc!2}\textbf{63.4}\dd{+1.6} & \multicolumn{1}{c}{--} & \cellcolor{lossc!32}33.7\dd{$-$22.8} & \multicolumn{1}{c}{--} \\
    \hline
    \multicolumn{6}{l}{\textit{Llama-3.1-8B-Instruct}} \\
        \ours{} (Ours) & 25 & 59.6 & \textbf{19.7} & 57.4 & \textbf{43.5} \\
        \quad w/o IA & 18 & \cellcolor{gainc!1}60.0\dd{+0.4} & 91.9 & \cellcolor{gainc!1}\textbf{57.8}\dd{+0.4} & 98.4 \\
        \quad w/o DG & 9 & \cellcolor{lossc!10}52.2\dd{$-$7.5} & 93.5 & \cellcolor{lossc!13}48.3\dd{$-$9.1} & 98.6 \\
        \quad w/o IA \& DG & 1 & \cellcolor{gainc!3}\textbf{61.7}\dd{+2.1} & \multicolumn{1}{c}{--} & \cellcolor{lossc!23}40.7\dd{$-$16.7} & \multicolumn{1}{c}{--} \\
    \hline
    \multicolumn{6}{l}{\textit{Llama-3-Meerkat-8B}} \\
        \ours{} (Ours) & 25 & 61.5 & \textbf{29.5} & \textbf{59.2} & \textbf{54.6} \\
        \quad w/o IA & 18 & 61.7\dd{+0.2} & 93.5 & \cellcolor{lossc!2}57.6\dd{$-$1.6} & 98.6 \\
        \quad w/o DG & 9 & \cellcolor{lossc!9}55.1\dd{$-$6.4} & 93.5 & \cellcolor{lossc!13}49.6\dd{$-$9.6} & 98.6 \\
        \quad w/o IA \& DG & 1 & \cellcolor{gainc!2}\textbf{63.0}\dd{+1.5} & \multicolumn{1}{c}{--} & \cellcolor{lossc!30}37.7\dd{$-$21.4} & \multicolumn{1}{c}{--} \\
    \hline
    \end{tabular}
    \caption{Our ablation study. Interv.: questions intervened on (\%); \#Fwd: forward passes per question. w/o IA \& DG: No intervention}
    \label{tab:ablation}
\end{table}

On \hard{}, \ours{} raises accuracy by 16.7--22.8 points for the three models at a cost of 1.5--2.1 points on clean MedQA questions (Table~\ref{tab:main_acc}), and for Llama-3.1-8B-Instruct it damages a similar share of correctly answered questions on MMLU professional medicine (Appendix~\ref{app:cleancost}).
\begin{figure}[t]
    \centering
    \includegraphics[width=\columnwidth]{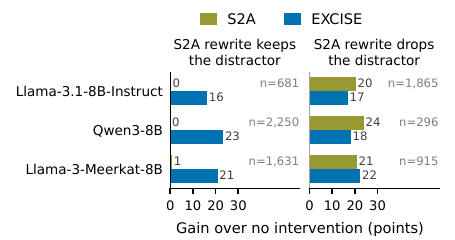}
    \caption{Gains of S2A and \ours{} on \hard{} (both distractor types), by whether the S2A rewrite keeps the distractor.}
    \label{fig:s2a_body}
\end{figure}
\paragraph{Against methods for the same task.}
On \hard{}, \ours{} has the highest accuracy among the methods without chain-of-thought generation in Table~\ref{tab:main_acc} for every model.
The ignore instruction and steering barely move \hard{} (at most +4.7 points), and with chain-of-thought generation, \ours{} is 14.2--22.0 points above CoT + I$^3$C on \hard{}.
S2A also matches or exceeds \ours{} for Llama-3.1-8B-Instruct on MedDistractQA and BiasMedQA.
Its gains depend on whether the model's rewrite leaves out the distractor, and where the rewrite keeps it, S2A gains at most 0.9 points on \hard{}, against 16--23 for \ours{} (Figure~\ref{fig:s2a_body}).
\ours{} is also faster, with 0.72--0.82 seconds per question against 3.2--4.2 for S2A, and with chain-of-thought generation 2.9--7.3 against 10.1--20.3 for CoT + I$^3$C.

\paragraph{Beyond \hard{}.}
On MedDistractQA and BiasMedQA, \ours{} gains 5.0--10.7 and 5.7--24.8 points, respectively
(Table~\ref{tab:main_acc}; per-bias results in Appendix~\ref{app:biasmedqa}), the best of these methods apart from S2A for Llama-3.1-8B-Instruct.
With chain-of-thought generation, it still gains 13.5--17.3 points on \hard{}, 3.5--7.0 on MedDistractQA and 4.3--16.5 on BiasMedQA, at a cost of 1.3--2.4 points on clean MedQA.
\section{Analysis}
\label{sec:analysis}
\subsection{Can \ours{} Find the Distractor among Added Noise?}
\label{sec:decoy}
\begin{figure}[t]
    \centering
    \includegraphics[width=\columnwidth]{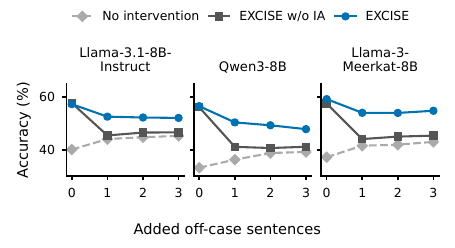}
    \caption{Accuracy on \hard{} (NL and BS pooled) with zero to three added off-case sentences.}
    \label{fig:decoy_k}
\end{figure}
\hard{} contains a single added distractor, but a case may also include other irrelevant remarks. We test whether \ours{} can locate the distractor among such noise by inserting one to three additional off-case sentences at random positions before the original distractor.
Each added sentence is the distractor of another question of the same type and shares no content word with any of this question's options, so it resembles the distractor in form but was not written to push the model toward any of them.
We compare against \ours{} w/o IA, a discordance-only variant that blocks the most discordant sentence that passes the gate.

Figure~\ref{fig:decoy_k} shows that the two methods perform similarly without additional remarks, but diverge once a remark is added.
With one added sentence, \ours{} outperforms \ours{} w/o IA by 7.2--10.0 accuracy points across three models, and the gap persists with further additions.
With one added sentence, \ours{} blocks the original distractor in 54--80\% of flipped questions (correct
on the original clean question but wrong on the augmented question), compared with 24--31\% for
\ours{} w/o IA.
The latter instead blocks the added remark in 59--67\%.
Although the added sentences weaken the distractor's influence, \ours{} remains more accurate because it selects the sentence whose blocking most weakens the current decision.
Further analysis is in Appendix~\ref{app:decoy}.

\subsection{What Does Each Part Contribute?}
\label{sec:selectivity}
\begin{figure}[t]
    \centering
    \includegraphics[width=\columnwidth]{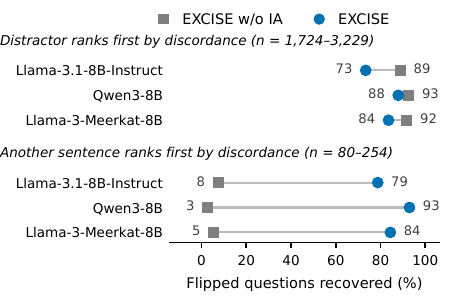}
    \caption{Flipped questions recovered by \ours{} w/o IA and \ours{}, pooled over \hard{}, MedDistractQA and BiasMedQA. Only questions in which the distractor passes the gate are included, grouped by whether the distractor or another sentence ranks first by discordance.}
    \label{fig:regime_body}
\end{figure}

The added-noise experiment highlights the role of attribution; we now examine how influence and discordance complement each other (Table~\ref{tab:ablation}).
Without the gate, blocking the top-attributed sentence in every question (\ours{} w/o DG) costs 7.9--10.2 points on clean questions relative to no intervention, because in questions without a distractor this sentence is often a key clinical finding that fits the case.
Without attribution, \ours{} w/o IA reaches accuracy comparable to \ours{}, because in these benchmarks the distractor is usually the most discordant sentence (Section~\ref{sec:conceptual}).
When another sentence is more discordant than the distractor, in 4.1--8.9\% of flipped questions, \ours{} recovers most of them and \ours{} w/o IA almost none, since \ours{} w/o IA blocks the least fitting sentence whether or not it moves the decision (Figure~\ref{fig:regime_body}).

Attribution also makes the decision to intervene informative.
\ours{} intervenes on 79--94\% of \hard{} questions whose answer the distractor pulls to the target, but on only 7--21\% of those answered correctly with and without it (Appendix~\ref{app:when}), whereas \ours{} w/o IA intervenes on 91.9--93.5\% of clean questions, so whether it intervenes says nothing about whether the model was misled.

We also compare each component with alternatives.
For Llama-3.1-8B-Instruct and Qwen3-8B, ContextCite under the same gate is within one point of IA but needs 3.3 times as many forward passes; CAMAB, averaged attention and AT2 fall at least 2.0 points behind on \hard{} (Appendix~\ref{app:top1}).
When the distractor is moved to the middle or start of the narrative, \ours{} changes by at most 3.2 points, against up to 13.4 for averaged attention under the same gate (Appendix~\ref{app:position}), and sentence encoders work worse than the model's own representations in the gate (Appendix~\ref{app:surface}).

\subsection{Where Does \ours{} Fail?}
\label{sec:funnel}

On flipped \hard{} questions, attribution ranks the distractor first in 90--91\%; the gate passes it in 81--95\% of these, and blocking it then restores the answer in 93--95\% (Appendix~\ref{app:when}).
Overall, \ours{} recovers 71--81\% of flipped questions, against 90--94\% with the known distractor blocked.
The gate is the weakest step for \hard{} Bystander distractors, which it passes in 81--86\% of questions, possibly because a relative's medical history is closer to the case than other distractors are.
Attribution is the weakest step on MedDistractQA, where the distractor ranks first in only 72--89\% of questions.

\section{Conclusion}
\label{sec:conclusion}
We presented \ours{}, which selects the sentence that most influences a medical QA model's answer and blocks attention to it only when the model's own hidden states mark it as inconsistent with the case.
Unlike generation-based defenses that ask the model to identify or rewrite irrelevant content, \ours{} makes these decisions from measured intervention effects and hidden-state discordance.

Our analysis showed that blocking a known distractor
largely restores accuracy, whereas fixed head masks
can suppress correct answers.
Hidden states also distinguish distractors even
when the model follows them.

On \hard{}, \ours{} improves accuracy by 16.7--22.8 points at a 1.5--2.1-point cost on clean questions, with gains extending to other benchmarks and chain-of-thought generation.
Our ablations show that attribution helps locate the original distractor among additional off-case remarks, while the discordance gate limits harmful interventions on clean questions.
Together, these signals make intervention selective: \ours{} intervenes far more often when a distractor draws the model toward its targeted wrong option than when the answer remains correct despite its presence.
\section*{Limitations}

\paragraph{Scope.}
\ours{} is designed to suppress information irrelevant to answering the question, rather than to correct false medical claims.
Fabricated rules, altered numerical values, and misleading guideline statements, such as those studied in MedMisBench \citep{zhou2026medmisbench}, require fact verification rather than ignoring and are outside its scope.
Our experiments focus on answer-preserving distractors in medical QA rather than the full range of noise encountered in clinical records.
Although Section~\ref{sec:decoy} introduces additional off-case remarks, these do not capture transcription errors, speech-recognition artifacts, or duplicated text.
Moreover, each question contains only one answer-targeting distractor; the additional remarks are not designed to favor wrong options.

\paragraph{Applicability.}
\ours{} requires access to answer-option logits, attention, and hidden states.
It therefore applies to multiple-choice questions with white-box model access.
Although its decisions carry over to chain-of-thought generation (Table~\ref{tab:main_acc}), questions without answer options are not covered.
The discordance gate also requires at least three narrative sentences; shorter questions are left unchanged.

The gate estimates a mean and first principal component from the unlabeled inputs to which it is applied.
This batch-dependent projection can be difficult to estimate when questions arrive individually or when suitable batches cannot be gathered.
A projection fixed on development data avoids this requirement, but retains a smaller accuracy gain (Appendix~\ref{app:basis}).

\paragraph{Cost and Trade-offs.}
\ours{} uses $3n+1$ forward passes per question: one answer readout, $n$ attribution passes, and $2n$ short passes for the discordance gate.
On \hard{}, this amounts to about 25 passes and 721--819\,ms per question, approximately 18--20 times the latency of a single pass.
The method also reduces accuracy on clean MedQA questions by 1.5--2.1 points (Table~\ref{tab:main_acc}), reflecting the remaining risk of unnecessary intervention.

\section*{Ethical Considerations}

\hard{} consists of synthetic distractors written by GPT-OSS-120B \citep{gptoss} for public MedQA questions and contains no patient data.
Available dataset and model licenses are listed in Appendix~\ref{app:licenses}.
\ours{} is an analysis and mitigation method for benchmark settings.
It is not validated for clinical use, and its clean cost and remaining errors mean that it must not be deployed for patient care without clinical evaluation.

\section*{Acknowledgments}
We thank Heeeun Jung and Nahyun Lim from the Epidemiology of Aging Lab, College of Medicine, Kyung Hee University, for volunteering to review the \hard{} questions.

\bibliography{latex/custom}
\appendix
\raggedbottom

\FloatBarrier
\section{Artifact Licenses}
\label{app:licenses}
\subsection{Datasets}
We use the following datasets:
\begin{itemize}
    \item MedQA, with MIT listed in the original repository and CC BY-SA 4.0 listed in the Hugging Face distribution we use (GBaker/MedQA-USMLE-4-options-hf\footnote{\url{https://huggingface.co/datasets/GBaker/MedQA-USMLE-4-options-hf}}).
    \item MedDistractQA, under CC BY 4.0.
    \item BiasMedQA, for which the source repository does not specify a dataset license; we use it for evaluation and do not redistribute it.
    \item MMLU professional medicine, under MIT.
\end{itemize}
\hard{} includes MedQA question text and answer options with distractors constructed by our pipeline.
Licensing and source attribution details are provided in the accompanying dataset card.\footnote{\url{https://huggingface.co/datasets/yrhong/MedQA-TD}}

\subsection{Models}
We use the following models:
\begin{itemize}
    \item Llama-3.1-8B-Instruct, 8B parameters, under the Llama 3.1 Community License.
    \item Qwen3-8B, 8.2B parameters, under the Apache 2.0 License.
    \item Llama-3-8B-Instruct, 8B parameters, under the Meta Llama 3 Community License.
    \item Granite-3.3-8B-Instruct, 8B parameters, under the Apache 2.0 License.
    \item Falcon3-10B-Instruct, 10B parameters, under the TII Falcon-LLM License 2.0.
    \item Phi-4, 14B parameters, under the MIT License.
    \item MedGemma-27B-text-it, 27B parameters, under the Health AI Developer Foundations Terms of Use.
    \item HuatuoGPT-o1-8B, 8B parameters, under the Apache 2.0 License.
    \item Llama-3-Meerkat-8B, 8B parameters, with a CC BY-NC 4.0 model-card license.
    \item OLMo-2-7B-Instruct, 7B parameters, under the Apache 2.0 License.
    \item GPT-OSS-120B, 117B parameters, under the Apache 2.0 License.
    \item SimCSE (sup-simcse-roberta-large), 355M parameters, under the MIT License.
    \item SapBERT (SapBERT-from-PubMedBERT-fulltext), 110M parameters, under the Apache 2.0 License.
    \item all-mpnet-base-v2, 110M parameters, under the Apache 2.0 License.
\end{itemize}

\section{Benchmark}
\label{app:sec_benchmark}
\begin{figure}[!t]
    \includegraphics[width=\columnwidth]{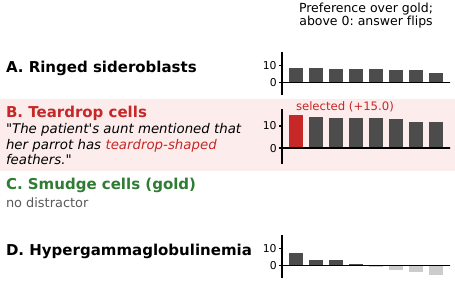}
    \caption{Constructing \hard{} for one MedQA question (gold: C). Each bar is one of the eight distractors written for a wrong option, with the reference model's preference for its strongest wrong option over the gold option. The flipping distractor with the largest margin is kept (red).}
    \label{fig:construction}
\end{figure}

Figure~\ref{fig:construction} illustrates how a distractor
is selected for \hard{}.

Table~\ref{tab:multimodel_full} gives the accuracy and target-wrong answers of the nine open models.
GPT-OSS-120B first writes eight candidates per wrong option.
We then ask GPT-OSS-120B which answer options each candidate points to, without solving the medical question (Appendix~\ref{app:prompts}).
The check is repeated with the options in ABCD and DCBA order, keeping their labels unchanged.
In both orders, the response must indicate an association with either the intended wrong option alone or multiple wrong options, without including the gold option.
Candidates are rejected if either response reports no association, identifies only a different wrong option, includes the gold option, or cannot be parsed.
Rejected Nonliteral candidates are regenerated up to twice per slot, leaving at most eight accepted candidates per
wrong option. Bystander candidates are not regenerated.

The reference model then answers the question with each accepted candidate, scoring A--D at the first answer token before generating a rationale.
If any candidate turns a uniquely correct clean prediction into a uniquely wrong prediction, we select among those candidates. Otherwise, we consider all accepted candidates.
We keep the candidate with the largest margin in Eq.~\ref{eq:selection}.
Exact ties are resolved by merging identical texts and then selecting the candidate with the smallest salted SHA-256 hash.

For questions whose wrong options offer no usable clinical topic (e.g., single letters or numbers), no candidate passes the check, and a fixed template replaces the generated distractor in 14 test cases, seven per type.

\begin{table*}[t]
    \centering
    \small
    \setlength{\tabcolsep}{4pt}
    \begin{tabular}{l c c c c c c c c c}
    \hline
    & \multicolumn{5}{c}{\textbf{Accuracy (\%)}} & \multicolumn{4}{c}{\textbf{Target-wrong answers (\%)}} \\
    \cmidrule(lr){2-6} \cmidrule(lr){7-10}
    & & \multicolumn{2}{c}{\textbf{MDQA}} & \multicolumn{2}{c}{\textbf{\hard{}}} & \multicolumn{2}{c}{\textbf{Nonliteral}} & \multicolumn{2}{c}{\textbf{Bystander}} \\
    \cmidrule(lr){3-4} \cmidrule(lr){5-6} \cmidrule(lr){7-8} \cmidrule(lr){9-10}
    \textbf{Model} & \textbf{Clean} & \textbf{NL} & \textbf{BS} & \textbf{NL} & \textbf{BS} & \textbf{Clean} & \textbf{\hard{}} & \textbf{Clean} & \textbf{\hard{}} \\
    \hline
    \multicolumn{10}{l}{\textit{General}} \\
    Llama-3.1-8B-Instruct & 61.7 & 53.1 & 55.9 & 39.2 & 42.1 & 17.4 & 45.8 & 17.8 & 42.6 \\
    Qwen3-8B & 63.4 & 49.4 & 51.1 & 36.4 & 31.0 & 19.0 & 53.0 & 17.7 & 59.5 \\
    Llama-3-8B-Instruct & 60.0 & 49.6 & 49.9 & 37.0 & 35.4 & 18.9 & 49.1 & 17.0 & 48.2 \\
    Granite-3.3-8B-Instruct & 51.5 & 40.1 & 38.6 & 30.6 & 27.5 & 20.9 & 52.2 & 19.0 & 57.1 \\
    Falcon3-10B-Instruct & 58.4 & 46.2 & 46.4 & 33.9 & 29.1 & 18.9 & 53.7 & 18.5 & 59.0 \\
    Phi-4 (14B) & 70.9 & 61.9 & 64.9 & 51.7 & 49.6 & 14.8 & 37.9 & 14.5 & 39.2 \\
    \hline
    \multicolumn{10}{l}{\textit{Medical}} \\
    MedGemma-27B-text-it & 73.3 & 64.7 & 65.6 & 52.2 & 51.5 & 12.5 & 38.3 & 11.9 & 38.2 \\
    HuatuoGPT-o1-8B$^\dagger$ & 74.6 & 66.5 & 64.5 & 54.8 & 55.4 & 10.3 & 31.3 & 10.0 & 31.3 \\
    \hline
    \multicolumn{10}{l}{\textit{Reasoning}} \\
    Qwen3-8B (thinking)$^\dagger$ & 77.5 & 69.8 & 70.3 & 58.3 & 55.8 & 10.5 & 34.5 & 10.4 & 37.3 \\
    \hline
    \end{tabular}
    \caption{The nine open models without intervention on the 1,273 test questions: Llama-3.1-8B-Instruct and Llama-3-8B-Instruct \citep{llama3}, Qwen3-8B \citep{qwen3}, Granite-3.3-8B-Instruct \citep{granite33}, Falcon3-10B-Instruct \citep{falcon3}, Phi-4 \citep{abdin2024phi4}, MedGemma-27B-text-it \citep{sellergren2025medgemma} and HuatuoGPT-o1-8B \citep{huatuo}. Accuracy: clean and with the MedDistractQA (MDQA) and \hard{} distractors of both types (NL, BS). Target-wrong answers: answers selecting the option the distractor was built to promote, on the clean questions and on \hard{}. $^\dagger$Reasons before answering and is read from its generated answer (unparsable outputs, at most 0.8\%, count as wrong).}
    \label{tab:multimodel_full}
\end{table*}

\subsection{Human Review of \hard{} Questions}
\label{app:review}
The two reviewers are PhD students in a geriatric epidemiology laboratory of a medical school, working on clinical cohort studies, and neither took part in building \hard{}.
Both reviewers volunteered.
They judged 100 test questions sampled at random, 50 of each distractor type, independently, and each answered two questions for each: whether the inserted sentence changes which option is correct, and whether a clinician would use it as evidence for the question.
Neither reviewer answered yes to either question.
For the second question, they answered possibly for 1 and 4 cases, with one case in common.
They agreed on all 100 cases for the first question and on 97 for the second (Gwet's AC1 0.97; \citealp{gwet2008computing}), and we do not report Cohen's kappa \citep{cohen1960coefficient}, which becomes uninformative when almost all answers fall into one category \citep{feinstein1990high}.

\subsection{What Makes a Distractor Strong}
\label{app:strength}
\begin{figure*}[t]
    \centering
    \includegraphics[width=\linewidth]{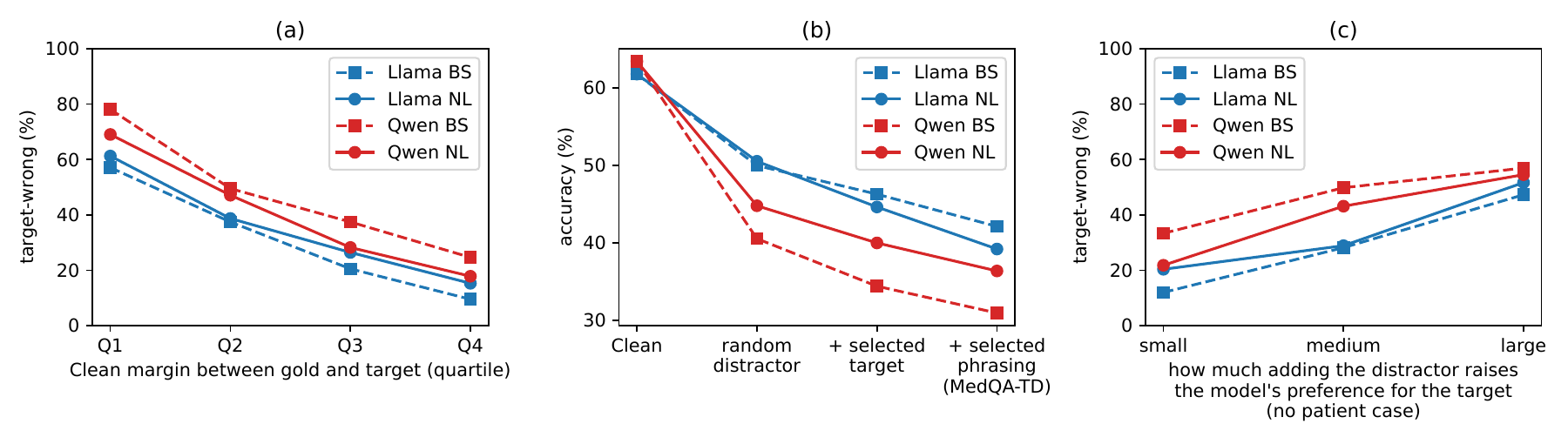}
    \caption{What makes \hard{} hard (Llama-3.1-8B-Instruct and Qwen3-8B). (a) Target-wrong rate among questions answered correctly without the distractor, by quartile of the clean margin between the gold and the targeted option. (b) Accuracy on all 1,273 questions with a random distractor (random target and phrasing), a random phrasing for the selected target, and the selected distractor (\hard{}). (c) Target-wrong rate on the questions of (a), by third of the case-free association: the increase in the target-minus-gold logit when the distractor is added to the question sentence and options alone.}
    \label{fig:vulnerability}
\end{figure*}

Figure~\ref{fig:vulnerability} analyzes what makes a distractor of \hard{} strong for Llama-3.1-8B-Instruct and Qwen3-8B.
First, distractors flip most often the questions on which the model was already torn between the gold and the targeted option before the distractor was added.
Among questions answered correctly without the distractor, 57--78\% of those in the quartile with the smallest clean margin between the two options flip to the target, against 10--25\% in the quartile with the largest margin (Figure~\ref{fig:vulnerability}(a)).
Second, the targeted option and the phrasing of the selected distractor each lower accuracy further than a distractor drawn at random from the same candidates (Figure~\ref{fig:vulnerability}(b)).
Each of the two steps lowers accuracy by 3.5--6.1 points for both models and both distractor types, with every 95\% interval below zero.
Third, the selected phrasings push toward the target even without the case (Figure~\ref{fig:vulnerability}(c)).

\subsection{Position of the Distractor}
\label{app:position}
\begin{table*}[!ht]
    \centering
    \footnotesize
    \setlength{\tabcolsep}{4pt}
    \begin{tabular}{l l l c c c}
    \hline
    \textbf{Position} & \textbf{Data} & \textbf{Model} & \textbf{\ours{}} & \textbf{Last} & \textbf{Known distractor} \\
    \hline
    End & \hard{} NL & Llama-3.1-8B-Instruct & +19.0 & +21.6 & +22.1 \\
     & \hard{} NL & Qwen3-8B & +22.3 & +25.3 & +25.9 \\
     & \hard{} BS & Llama-3.1-8B-Instruct & +14.5 & +18.6 & +19.1 \\
     & \hard{} BS & Qwen3-8B & +23.3 & +30.9 & +31.7 \\
 & \hard{} NL & Llama-3-Meerkat-8B & +23.3 & +25.3 & +25.8 \\
 & \hard{} BS & Llama-3-Meerkat-8B & +19.6 & +24.5 & +25.4 \\
    \hline
    Middle & \hard{} NL, moved & Llama-3.1-8B-Instruct & +11.2 & $-$8.2 & +14.2 \\
     & \hard{} NL, moved & Qwen3-8B & +15.6 & $-$8.8 & +18.5 \\
     & \hard{} BS, moved & Llama-3.1-8B-Instruct & +6.8 & $-$7.5 & +11.0 \\
     & \hard{} BS, moved & Qwen3-8B & +16.4 & $-$9.1 & +23.4 \\
 & \hard{} NL, moved & Llama-3-Meerkat-8B & +10.4 & $-$8.5 & +14.0 \\
 & \hard{} BS, moved & Llama-3-Meerkat-8B & +10.3 & $-$7.5 & +15.0 \\
     & BiasMedQA conf. & Llama-3.1-8B-Instruct & $-$1.3 & $-$3.1 & $-$1.4 \\
     & BiasMedQA conf. & Qwen3-8B & +14.8 & $-$8.6 & +17.0 \\
 & BiasMedQA conf. & Llama-3-Meerkat-8B & $-$1.0 & $-$6.4 & +0.5 \\
    \hline
    First & \hard{} NL, first & Llama-3.1-8B-Instruct & +12.2 & $-$5.3 & +14.1 \\
     & \hard{} NL, first & Qwen3-8B & +13.9 & $-$6.8 & +16.3 \\
     & \hard{} BS, first & Llama-3.1-8B-Instruct & +8.0 & $-$5.3 & +11.5 \\
     & \hard{} BS, first & Qwen3-8B & +15.9 & $-$6.5 & +21.8 \\
 & \hard{} NL, first & Llama-3-Meerkat-8B & +9.9 & $-$5.4 & +12.4 \\
 & \hard{} BS, first & Llama-3-Meerkat-8B & +10.5 & $-$4.9 & +14.9 \\
    \hline
    \end{tabular}
    \caption{Accuracy change (points) by the position of the distractor, on held-out data. End: \hard{} as built, where the distractor is the last narrative sentence in 97\% of questions; Middle: the same test questions with the distractor moved to the middle of the narrative, and BiasMedQA \citep{schmidgall2024biasmedqa} confirmation, whose cue is the second narrative sentence; First: the same test questions with the distractor as the first sentence. Last: block the last narrative sentence in every question; Known distractor: block the distractor itself; both with the same eligibility, blocking and full denominators as \ours{}. The End row of Known distractor is therefore slightly below the gains of Section~\ref{sec:diagnosis}, which block in every question.}
    \label{tab:position}
\end{table*}
To analyze position bias, \hard{} comes in three versions, with the distractor at the end of the narrative (as built, where it is the last narrative sentence in 97\% of questions), in its middle or as its first sentence.
In the middle and first versions, \ours{} still gains 6.8--16.4 points, whereas blocking the last narrative sentence lowers accuracy by 4.9--9.1 points (Table~\ref{tab:position}).
Where a benchmark itself places the cue early in the case, as the second narrative sentence (BiasMedQA confirmation), \ours{} gains 14.8 points for Qwen3-8B and is within 1.3 points of no intervention for Llama-3.1-8B-Instruct and Llama-3-Meerkat-8B.
Table~\ref{tab:position_methods} gives every method at the three positions. The accuracy of \ours{} varies by at most 3.2 points across positions, so its gain is smaller where the distractor is less harmful; averaged attention with the gate varies by up to 13.4 points, falling from the end to the first position for Llama-3.1-8B-Instruct and Llama-3-Meerkat-8B and rising for Qwen3-8B, and ContextCite with the gate stays within 1.6 points of \ours{} at every position.
Selecting the sentence by its effect under blocking, as \ours{} and ContextCite do, is therefore insensitive to its position, unlike selecting it by attention.
Without intervention, every one of the nine models is least accurate with the distractor at the end, 4.1--10.5 points below the middle or first position (Table~\ref{tab:position_models}).

\begin{table*}[tp]
    \centering
    \small
    \setlength{\tabcolsep}{4pt}
    \begin{tabular}{l cccc cccc}
    \hline
    \textbf{Method} & \multicolumn{4}{c}{\textbf{\hard{} NL}} & \multicolumn{4}{c}{\textbf{\hard{} BS}} \\
    \cmidrule(lr){2-5} \cmidrule(lr){6-9}
    & End & Middle & First & Mean & End & Middle & First & Mean \\
    \hline
    \multicolumn{9}{l}{\textit{Qwen3-8B}} \\
        No intervention & 36.4 & 44.7 & 46.9 & 42.7 & 30.9 & 39.5 & 41.5 & 37.3 \\
        Ignore instruction & 35.4 & 44.9 & 46.8 & 42.4 & 31.0 & 39.7 & 41.6 & 37.4 \\
        CAA & 40.5 & 47.2 & 49.6 & 45.7 & 36.2 & 44.1 & 43.0 & 41.1 \\
        Avg.\ Attention (top-1) & 33.9 & 42.1 & 47.8 & 41.3 & 26.7 & 34.2 & 41.2 & 34.0 \\
        AT2 (top-1) & -- & -- & -- & -- & -- & -- & -- & -- \\
        CAMAB (top-1) & 46.4 & 46.0 & 47.8 & 46.7 & 47.3 & 45.9 & 46.8 & 46.7 \\
        ContextCite (top-1) & 47.5 & 45.9 & 47.5 & 46.9 & 48.2 & 45.1 & 46.6 & 46.6 \\
        \ours{} w/o DG & 47.5 & 45.2 & 46.9 & 46.5 & 46.9 & 44.5 & 46.4 & 46.0 \\
        \ours{} w/o IA & 59.5 & 61.2 & 61.8 & 60.8 & 52.5 & 55.0 & 57.0 & 54.9 \\
        Avg.\ Attention + DG & 37.6 & 45.6 & 51.1 & 44.7 & 31.0 & 39.0 & 43.3 & 37.8 \\
        AT2 + DG & -- & -- & -- & -- & -- & -- & -- & -- \\
        CAMAB + DG & 54.8 & 56.6 & 57.7 & 56.3 & 51.5 & 53.0 & 53.3 & 52.6 \\
        ContextCite + DG & 58.2 & 60.1 & 60.4 & 59.6 & 54.8 & 55.7 & 55.9 & 55.5 \\
        \ours{} & 58.7 & 60.3 & 60.8 & 59.9 & 54.2 & 55.9 & 57.4 & 55.9 \\
    \hline
    \multicolumn{9}{l}{\textit{Llama-3.1-8B-Instruct}} \\
        No intervention & 39.2 & 47.0 & 47.4 & 44.5 & 42.1 & 49.6 & 49.2 & 47.0 \\
        Ignore instruction & 39.4 & 47.5 & 46.4 & 44.4 & 41.6 & 49.4 & 47.8 & 46.2 \\
        CAA & 40.6 & 49.6 & 48.6 & 46.3 & 42.9 & 50.2 & 50.0 & 47.7 \\
        Avg.\ Attention (top-1) & 54.4 & 44.5 & 45.2 & 48.0 & 48.9 & 43.3 & 41.8 & 44.6 \\
        AT2 (top-1) & 51.1 & 47.7 & 47.4 & 48.7 & 45.4 & 45.2 & 45.7 & 45.5 \\
        CAMAB (top-1) & 47.9 & 47.8 & 49.6 & 48.4 & 47.5 & 47.0 & 48.0 & 47.5 \\
        ContextCite (top-1) & 48.1 & 48.1 & 48.8 & 48.3 & 48.5 & 46.9 & 47.0 & 47.5 \\
        \ours{} w/o DG & 48.1 & 46.7 & 48.5 & 47.8 & 48.4 & 46.0 & 47.5 & 47.3 \\
        \ours{} w/o IA & 59.5 & 59.7 & 59.8 & 59.6 & 56.2 & 56.2 & 56.9 & 56.4 \\
        Avg.\ Attention + DG & 55.9 & 51.6 & 49.6 & 52.4 & 51.9 & 50.4 & 49.3 & 50.5 \\
        AT2 + DG & 54.0 & 52.2 & 50.5 & 52.2 & 48.5 & 50.8 & 50.8 & 50.0 \\
        CAMAB + DG & 55.5 & 56.5 & 57.0 & 56.3 & 54.5 & 55.2 & 56.2 & 55.3 \\
        ContextCite + DG & 57.9 & 58.6 & 58.4 & 58.3 & 55.7 & 56.5 & 56.8 & 56.3 \\
        \ours{} & 58.2 & 58.3 & 59.5 & 58.7 & 56.6 & 56.4 & 57.2 & 56.7 \\
    \hline
    \multicolumn{9}{l}{\textit{Llama-3-Meerkat-8B}} \\
        No intervention & 37.5 & 49.1 & 50.1 & 45.6 & 38.0 & 48.3 & 47.5 & 44.6 \\
        Ignore instruction & 37.5 & 49.0 & 50.8 & 45.7 & 37.7 & 47.7 & 47.2 & 44.2 \\
        CAA & 38.5 & 51.1 & 51.8 & 47.1 & 40.9 & 51.1 & 49.9 & 47.3 \\
        Avg.\ Attention (top-1) & 57.3 & 48.5 & 47.9 & 51.2 & 52.1 & 43.5 & 42.8 & 46.1 \\
        AT2 (top-1) & -- & -- & -- & -- & -- & -- & -- & -- \\
        CAMAB (top-1) & 51.2 & 48.4 & 50.3 & 50.0 & 49.6 & 49.6 & 50.4 & 49.9 \\
        ContextCite (top-1) & 51.1 & 47.7 & 49.2 & 49.4 & 50.7 & 48.5 & 48.8 & 49.3 \\
        \ours{} w/o DG & 50.8 & 47.0 & 48.7 & 48.8 & 48.5 & 47.9 & 48.2 & 48.2 \\
        \ours{} w/o IA & 59.8 & 61.1 & 61.2 & 60.7 & 55.4 & 58.8 & 58.4 & 57.5 \\
        Avg.\ Attention + DG & 60.1 & 55.6 & 52.2 & 56.0 & 53.4 & 50.2 & 48.4 & 50.7 \\
        AT2 + DG & -- & -- & -- & -- & -- & -- & -- & -- \\
        CAMAB + DG & 58.1 & 58.8 & 59.3 & 58.8 & 56.6 & 57.5 & 58.0 & 57.3 \\
        ContextCite + DG & 60.2 & 60.9 & 60.1 & 60.4 & 58.8 & 58.6 & 58.6 & 58.7 \\
        \ours{} & 60.7 & 59.5 & 60.0 & 60.1 & 57.7 & 58.6 & 58.0 & 58.1 \\
    \hline
    \end{tabular}
    \caption{Accuracy (\%) with the distractor at the three positions of Table~\ref{tab:position} (End: \hard{} as built, as in the main tables; Middle; First) and their mean, for the first-token methods of Table~\ref{tab:main_acc}, the top-1 attribution baselines of Table~\ref{tab:top1} and the variants of Table~\ref{tab:ablation}. S2A and the chain-of-thought rows are not run at the other positions. The same 1,273 test questions and the same sentence are used in each position. AT2 is Llama-3.1-8B-Instruct only (--).}
    \label{tab:position_methods}
\end{table*}

\begin{table*}[tp]
    \centering
    \small
    \setlength{\tabcolsep}{4pt}
    \begin{tabular}{l cccc cccc}
    \hline
    \textbf{Model} & \multicolumn{4}{c}{\textbf{\hard{} NL}} & \multicolumn{4}{c}{\textbf{\hard{} BS}} \\
    \cmidrule(lr){2-5} \cmidrule(lr){6-9}
    & End & Middle & First & Mean & End & Middle & First & Mean \\
    \hline
    Llama-3.1-8B-Instruct & 39.2 & 47.0 & 47.4 & 44.5 & 42.1 & 49.6 & 49.2 & 47.0 \\
    Qwen3-8B & 36.4 & 44.7 & 46.9 & 42.7 & 30.9 & 39.5 & 41.5 & 37.3 \\
    Llama-3-8B-Instruct & 37.0 & 44.0 & 44.4 & 41.8 & 35.4 & 45.1 & 44.9 & 41.8 \\
    Granite-3.3-8B-Instruct & 30.6 & 37.4 & 36.3 & 34.8 & 27.5 & 34.5 & 31.6 & 31.2 \\
    Falcon3-10B-Instruct & 33.9 & 40.8 & 40.7 & 38.4 & 29.1 & 36.8 & 35.7 & 33.9 \\
    Phi-4 & 51.7 & 57.5 & 57.3 & 55.5 & 49.6 & 56.6 & 56.5 & 54.2 \\
    MedGemma-27B-text-it & 52.2 & 59.5 & 58.1 & 56.6 & 51.5 & 58.4 & 57.3 & 55.7 \\
    HuatuoGPT-o1-8B & 54.8 & 61.0 & 64.0 & 59.9 & 55.4 & 62.7 & 64.3 & 60.8 \\
    Qwen3-8B (thinking)$^\dagger$ & 58.3 & 64.3 & 67.5 & 63.3 & 55.8 & 61.4 & 64.8 & 60.6 \\
    \hline
    \end{tabular}
    \caption{Accuracy (\%) of the nine open models without intervention, with the distractor at the three positions (End: \hard{} as built). $^\dagger$Read from its generated answer, as in Table~\ref{tab:multimodel_full}.}
    \label{tab:position_models}
\end{table*}

\section{Experimental Details}
\label{app:sec_setup}

\label{app:data}
Table~\ref{tab:data} lists the development and test sets.
The development set is the whole MedQA validation split (1,269 usable of its 1,272 questions, Table~\ref{tab:data}), with \hard{} distractors built by the same pipeline as for the test questions, and it shares no question with the test set.
Development questions with no accepted candidate are excluded rather than assigned a fixed-template distractor.
\ours{} chooses its hyperparameters on all of it.
A distractor adds one narrative sentence and 16--20 words to a case of about seven sentences, and 93--99\% of these questions have at least three narrative sentences.
\begin{table*}[t]
    \centering
    \small
    \setlength{\tabcolsep}{3pt}
    \begin{tabular}{l l r c c c c}
    \hline
    \textbf{Set} & \textbf{Source} & \textbf{Questions} & \textbf{Options} & \textbf{Sentences} & \textbf{Words} & \textbf{$\geq$3 (\%)} \\
    \hline
    \multicolumn{7}{l}{\textit{Development}} \\
    Clean & MedQA val.\ \citep{MedQA} & 1,269 & 4 & 7.0 & 97.7 & 94.1 \\
    \hard{} Nonliteral & as above & 1,266 & 4 & 7.9 & 117.6 & 98.7 \\
    \hard{} Bystander & as above & 1,267 & 4 & 8.0 & 114.1 & 98.6 \\
    \hline
    \multicolumn{7}{l}{\textit{Test}} \\
    Clean & MedQA test \citep{MedQA} & 1,273 & 4 & 7.1 & 100.2 & 93.5 \\
    \hard{} Nonliteral & as above & 1,273 & 4 & 8.1 & 120.2 & 98.6 \\
    \hard{} Bystander & as above & 1,273 & 4 & 8.1 & 116.6 & 98.6 \\
    MedDistractQA NL & \citet{meddistractqa} & 1,273 & 4 & 8.1 & 118.9 & 98.5 \\
    MedDistractQA BS & \citet{meddistractqa} & 1,273 & 4 & 8.1 & 117.9 & 98.6 \\
    \hline
    \multicolumn{7}{l}{\textit{External}} \\
    BiasMedQA, clean & \citet{schmidgall2024biasmedqa} & 1,273 & 5 & 7.1 & 100.1 & 93.5 \\
    BiasMedQA, 7 types & \citet{schmidgall2024biasmedqa} & 7 $\times$ 1,273 & 5 & 8.0--8.1 & 113.6--120.1 & 98.3--98.6 \\
    MMLU prof.\ medicine & \citet{hendrycks2021mmlu} & 272 & 4 & 6.6 & 90.5 & 94.1 \\
    \hline
    \end{tabular}
    \caption{Development and test sets. Sentences: mean number of narrative sentences (the case before the question sentence); Words: mean number of words in them; $\geq$3: share of questions with at least three narrative sentences, the only ones on which \ours{} can intervene.}
    \label{tab:data}
\end{table*}

\subsection{Evaluation Protocol}
\label{app:inference}
We score each answer by the logits of the first answer token over the answer options (A--E for BiasMedQA), which measures the decision itself and makes attribution deterministic.
In chain-of-thought generation, \ours{} blocks the sentence chosen with this readout while the model writes its reasoning, and the prompts are in Appendix~\ref{app:prompts}.
All models run in bfloat16 with eager attention, one question at a time and without a key-value cache unless stated otherwise.
Blocking with strength $\atten$ is implemented by adding $\log\atten$ to the pre-softmax attention scores ($-\infty$ for $\atten = 0$), which is equivalent to the reweighting in Section~\ref{sec:method}.
The seconds per question in Table~\ref{tab:main_acc} are the mean over 50 \hard{} Nonliteral questions with batch size 1. Ties in the attribution score $\attr_i$, which bfloat16 logits make possible, go to the earlier sentence.
Deleting the selected sentence from the input instead of blocking it gives the same accuracy for Llama-3.1-8B-Instruct (+0.2 points [$-$0.3, +0.6] on clean questions and +0.6 [$-$0.3, +1.4] on \hard{}).

\subsection{Computational Budget and Infrastructure}
\label{app:budget}
Experiments were primarily conducted on NVIDIA RTX PRO 6000
Blackwell (96 GB) GPUs for evaluation and NVIDIA RTX A5000
(24 GB) GPUs for mechanistic analysis.
GPU-hours were calculated as the number of allocated GPUs
multiplied by execution time, excluding queue time.
Across all traceable experiments, including exploratory
and repeated runs, we estimate approximately 1,390 GPU-hours
(conservative lower bound: 1,340).
Of these, approximately 650 GPU-hours correspond to runs
underlying the reported results.

\subsection{Hyperparameters}
\label{app:hparams}
Every model is searched on the development set in the same way, maximizing the mean \hard{} accuracy with MedQA accuracy at most 0.5 points below that of a common starting setting (layer 16, $\thr = 1$, $\atten = 0.03$): the layer whose hidden states define the gate (middle layers: 8--24, 8--28 for Qwen3-8B, 10--30 for Phi-4), the gate threshold $\thr$ ($0.5$--$1.5$) and the blocking strength $\atten \in \{0, 0.01, 0.03, 0.1\}$.

Table~\ref{tab:hparams} lists the hyperparameters chosen for each model (Section~\ref{sec:method}) and the development accuracy they reach.
\begin{table*}[t]
    \centering
    \small
    \setlength{\tabcolsep}{4pt}
    \begin{tabular}{l c c c cc cc cc}
    \hline
    & & & & \multicolumn{2}{c}{\textbf{Clean}} & \multicolumn{4}{c}{\textbf{\hard{}}} \\
    \cmidrule(lr){5-6} \cmidrule(lr){7-10}
    & & & & & & \multicolumn{2}{c}{\textbf{NL}} & \multicolumn{2}{c}{\textbf{BS}} \\
    \cmidrule(lr){7-8} \cmidrule(lr){9-10}
    \textbf{Model} & \textbf{Layer} & $\thr$ & $\atten$ & None & \ours{} & None & \ours{} & None & \ours{} \\
    \hline
    Llama-3.1-8B-Instruct & 9 & 1 & 0 & 68.40 & 67.06 & 42.89 & 65.24 & 45.46 & 63.38 \\
    Qwen3-8B & 25 & 1 & 0 & 63.51 & 61.62 & 34.44 & 60.03 & 28.41 & 54.70 \\
    Llama-3-Meerkat-8B & 19 & 0.5 & 0 & 67.85 & 65.17 & 37.20 & 64.14 & 37.88 & 62.19 \\
    Phi-4 & 16 & 1 & 0.01 & 70.06 & 68.40 & 51.34 & 67.22 & 49.88 & 64.80 \\
    OLMo-2-7B-Instruct & 22 & 0.75 & 0 & 40.82 & 40.50 & 20.70 & 28.91 & 16.97 & 26.36 \\
    \hline
    \end{tabular}
    \caption{Hyperparameters of \ours{} chosen on the development set for each model, and development accuracy (\%) without intervention (None) and with \ours{}. $\thr$: gate threshold; $\atten$: blocking strength.}
    \label{tab:hparams}
\end{table*}

\subsection{Baseline Details}
\label{app:baselines}
\textbf{Averaged attention.} The score of a sentence is the attention probability that the answer position, where the option scores are read, assigns to the sentence's tokens, summed over these tokens and averaged over all heads and layers, without length normalization, as in the attention baselines of \citet{cohenwang2024contextcite}, \citet{liu2025attribot} and \citet{cohenwang2025at2}; the highest-scoring sentence is blocked.
\textbf{CAA.} The steering vector is the mean difference between the hidden states at the answer position of the clean and the \hard{} prompts (both distractor types) of 190 validation questions; it is added at one layer to every position of the answer prefix. The layer and strength are chosen on these questions by the errors recovered beyond a norm-matched random vector (Llama: layer 15, strength 2.0; Qwen: 23, 3.0; Meerkat: 15, 3.0).
\textbf{Other baselines.} CAMAB (40 queries), ContextCite (64 ablations) and AT2 (released estimator) keep the settings of their papers or released code.

\subsection{Statistical Tests}
\label{app:stats}
Table~\ref{tab:stats} gives the paired comparisons behind Sections~\ref{sec:results} and~\ref{sec:selectivity}.
Against the ignore instruction and CAA, every \hard{} difference is at least +13.7 points, and against S2A at least +3.5 points except for Llama-3.1-8B-Instruct on Nonliteral ($-$0.1); all comparisons and exact McNemar $p$-values are in the released results.
\begin{table*}[t]
    \centering
    \small
    \begin{tabular}{l c c c}
    \hline
    \textbf{\ours{} minus} & \textbf{Clean} & \textbf{\hard{} NL} & \textbf{\hard{} BS} \\
    \hline
    \multicolumn{4}{l}{\textbf{Qwen3-8B}} \\
    No intervention & $-$1.6 [$-$3.0, $-$0.3] & +22.3 [+19.9, +24.7] & +23.3 [+20.7, +25.8] \\
    \cdashline{1-4}
    Avg.\ Attention + DG & $-$1.1 [$-$2.4, +0.2] & +21.1 [+18.7, +23.4] & +23.2 [+20.7, +25.7] \\
    CAMAB + DG & $-$0.5 [$-$1.5, +0.5] & +3.9 [+2.6, +5.3] & +2.7 [+1.5, +4.0] \\
    ContextCite + DG & $-$0.5 [$-$1.4, +0.3] & +0.5 [$-$0.3, +1.3] & $-$0.6 [$-$1.6, +0.2] \\
    \cdashline{1-4}
    \ours{} w/o DG & +8.6 [+6.0, +11.2] & +11.2 [+9.0, +13.6] & +7.3 [+4.9, +9.8] \\
    \ours{} w/o IA & +0.1 [$-$1.2, +1.3] & $-$0.8 [$-$1.9, +0.3] & +1.6 [+0.1, +3.2] \\
    \hline
    \multicolumn{4}{l}{\textbf{Llama-3.1-8B-Instruct}} \\
    No intervention & $-$2.1 [$-$3.5, $-$0.8] & +19.0 [+16.6, +21.4] & +14.5 [+12.2, +16.7] \\
    \cdashline{1-4}
    Avg.\ Attention + DG & $-$1.4 [$-$2.7, $-$0.2] & +2.3 [+0.9, +3.7] & +4.7 [+2.9, +6.5] \\
    AT2 + DG & $-$1.1 [$-$2.4, +0.2] & +4.2 [+2.7, +5.9] & +8.1 [+6.1, +10.1] \\
    CAMAB + DG & $-$0.6 [$-$1.5, +0.2] & +2.7 [+1.3, +4.2] & +2.0 [+0.7, +3.5] \\
    ContextCite + DG & 0.0 [$-$0.7, +0.7] & +0.3 [$-$0.6, +1.3] & +0.9 [$-$0.2, +2.0] \\
    \cdashline{1-4}
    \ours{} w/o DG & +7.5 [+4.9, +10.1] & +10.1 [+7.8, +12.5] & +8.2 [+5.7, +10.6] \\
    \ours{} w/o IA & $-$0.4 [$-$1.6, +0.8] & $-$1.3 [$-$2.4, $-$0.1] & +0.4 [$-$0.9, +1.7] \\
    \hline
    \multicolumn{4}{l}{\textbf{Llama-3-Meerkat-8B}} \\
    No intervention & $-$1.5 [$-$3.2, +0.2] & +23.3 [+20.7, +26.0] & +19.6 [+17.0, +22.2] \\
    \cdashline{1-4}
    Avg.\ Attention + DG & $-$1.0 [$-$2.5, +0.4] & +0.6 [$-$0.8, +2.0] & +4.2 [+2.4, +6.0] \\
    CAMAB + DG & $-$1.1 [$-$2.3, +0.1] & +2.6 [+1.2, +4.1] & +1.1 [$-$0.2, +2.4] \\
    ContextCite + DG & $-$0.2 [$-$1.3, +0.8] & +0.5 [$-$0.5, +1.5] & $-$1.1 [$-$2.2, $-$0.1] \\
    \cdashline{1-4}
    \ours{} w/o DG & +6.4 [+4.0, +8.9] & +10.0 [+8.0, +12.0] & +9.2 [+7.0, +11.5] \\
    \ours{} w/o IA & $-$0.2 [$-$1.8, +1.3] & +0.9 [$-$0.5, +2.4] & +2.3 [+0.6, +4.0] \\
    \hline
    \end{tabular}
    \caption{Paired comparisons on the test sets: accuracy of \ours{} minus that of each method (points), with 95\% bootstrap intervals over the 1,273 questions (10,000 resamples). Dashed rules separate no intervention, attribution methods with the Discordance Gate (DG), and single modules; AT2 is Llama-3.1-8B-Instruct only.}
    \label{tab:stats}
\end{table*}

\section{Diagnosis}
\label{app:sec_diagnosis}

\subsection{Discordance of the Distractor}
\label{app:recognition}
We measure how well each narrative sentence $s_i$ fits the rest of the case with the discordance $\disc_i$ defined in Section~\ref{sec:method_dg}, which is the cosine similarity between the hidden states of $s_i$ read alone and of the rest of the narrative read alone, after removing the mean and the first principal component of the unlabeled batch, negated and standardized within the question.
We read the hidden states at the layer that \ours{} uses for each model (layer 9 for Llama-3.1-8B-Instruct, 25 for Qwen3-8B, 19 for Llama-3-Meerkat-8B, 16 for Phi-4 and 22 for OLMo-2-7B-Instruct).
Figure~\ref{fig:recognition_body} counts how often the distractor has the highest discordance among the narrative sentences, among the questions that the model answers correctly without the distractor and wrongly with it, and chance is the mean of one over the number of narrative sentences.

At these layers the first principal component carries 94--98\% of the variance for four of the five models but 9\% for OLMo-2-7B-Instruct.
For OLMo-2-7B-Instruct, removing it lowers the rate on \hard{} from 63\% to 29\% for Nonliteral and from 40\% to 20\% for Bystander, so in this model the removed component also carries differences between sentences.
We keep the setting chosen on the development set for this model.
On the test questions with the distractor moved to the middle of the narrative, the score ranks it first in 79--92\% of \hard{} questions for Llama-3.1-8B-Instruct and 70--90\% for Qwen3-8B, against 5--6\% without removing the first principal component.

To see how common such a dominant direction is, we encode the sentences of all 2,546 \hard{} test questions, and the rest of each narrative, with eleven open models at five depths (from a quarter to three quarters of the layers).
In nine models the first principal component carries at least half of the variance at every depth, and in eight of them (Llama-3.1-8B-Instruct, Meta-Llama-3-8B-Instruct, Llama-3-Meerkat-8B, HuatuoGPT-o1-8B, Qwen3-8B, Granite-3.3-8B-Instruct, Falcon3-10B-Instruct and Phi-4) removing it raises the rate at which the distractor ranks first from 1--7\% to 37--88\%.
In the ninth, MedGemma-27B-text-it, the rate stays at 2--17\% with or without removal.
Only the two OLMo-2 models lack such a direction, with the first component at 8--9\% of the variance for OLMo-2-7B-Instruct and at most 48\% for OLMo-2-32B-Instruct.

\subsection{Head Masks}
\label{app:headmask}
We follow the differentiable head masking of \citet{niu-etal-2025-llama} in the sentence-level form of \citet{liu2026sentence}.
A continuous mask on every attention head is trained to lower the target-minus-gold logit gap on \hard{} pairs, with a KL term that keeps the output on the clean question and a sparsity weight $\lambda$ that keeps few heads removed.

The masks are trained on the Bystander and Nonliteral pairs of 382 development questions, using only the pairs that the unmasked model answers correctly without the distractor and with the targeted option with it. 
We search the learning rate in $\{0.03, 0.1, 0.3\}$ and $\lambda$ in $\{1, 2\}$, train for 200 epochs, keep a mask every 10 epochs and then every epoch around the best one, which gives up to 130 candidates per model.
We pick the candidate with the highest mean of \hard{} Bystander and Nonliteral accuracy on the other 887 development questions among those whose clean accuracy stays within 2 points of the unmasked model.
Only OLMo-2-7B-Instruct had such a candidate, so for the other four models we used the one with the highest mean of this \hard{} accuracy and clean accuracy.
The gold-cue control uses the \hard{} construction with the distractor pointing at the correct option instead of a wrong one (same generator, judge, insertion rule and questions; one candidate drawn at random among those that evoke only the correct option).
For each tuned mask, five control masks remove the same number of heads from each layer, chosen uniformly at random within the layer, and are read on every set.
Table~\ref{tab:headmask} gives the accuracy of every mask on each set. In all five models, the selected masks lie above all five random masks on clean and \hard{} questions. On the gold-cue control they lie below all five in nine of the ten model and distractor-type cells, and within the random range for Llama-3-Meerkat-8B Nonliteral.
\begin{table*}[t]
    \centering
    \small
    \setlength{\tabcolsep}{4pt}
    \begin{tabular}{l c c c c c c}
    \hline
    & & \textbf{Clean} & \multicolumn{2}{c}{\textbf{\hard}} & \multicolumn{2}{c}{\textbf{Gold-cue control}} \\
    \cmidrule(lr){3-3} \cmidrule(lr){4-5} \cmidrule(lr){6-7}
    \textbf{Method} & \textbf{Heads} & \textbf{MedQA} & \textbf{NL} & \textbf{BS} & \textbf{NL} & \textbf{BS} \\
    \hline
        \multicolumn{7}{l}{\textbf{Qwen3-8B}} \\
    \quad No mask & -- & 64.65 & 37.39 & 32.68 & 81.07 & 85.07 \\
    \quad Head mask, tuned on development data & 126 & 54.83 & 55.15 & 55.38 & 38.18 & 34.17 \\
    \quad Random heads, same layers (mean of 5) & 126 & 50.95 & 26.35 & 22.70 & 76.81 & 80.97 \\
    \cdashline{1-7}
    \quad No intervention & -- & 63.39 & 36.37 & 30.95 & 82.17 & 86.33 \\
    \quad \ours{} & -- & 61.74 & 58.68 & 54.20 & 64.57 & 67.87 \\
    \hline
    \multicolumn{7}{l}{\textbf{Llama-3.1-8B-Instruct}} \\
    \quad No mask & -- & 61.74 & 39.04 & 41.32 & 73.68 & 68.97 \\
    \quad Head mask, tuned on development data & 83 & 55.77 & 58.84 & 58.84 & 25.61 & 19.01 \\
    \quad Random heads, same layers (mean of 5) & 83 & 40.42 & 27.64 & 25.01 & 55.44 & 59.86 \\
    \cdashline{1-7}
    \quad No intervention & -- & 61.74 & 39.20 & 42.11 & 73.84 & 69.84 \\
    \quad \ours{} & -- & 59.62 & 58.21 & 56.56 & 62.61 & 62.61 \\
    \hline
    \multicolumn{7}{l}{\textbf{Llama-3-Meerkat-8B}} \\
    \quad No mask & -- & 62.69 & 38.41 & 36.61 & 78.87 & 79.10 \\
    \quad Head mask, tuned on development data & 83 & 56.72 & 57.11 & 55.07 & 37.08 & 35.27 \\
    \quad Random heads, same layers (mean of 5) & 83 & 35.52 & 23.11 & 20.33 & 51.66 & 56.26 \\
    \cdashline{1-7}
    \quad No intervention & -- & 63.00 & 37.47 & 38.02 & 79.50 & 80.20 \\
    \quad \ours{} & -- & 61.51 & 60.72 & 57.66 & 64.73 & 65.44 \\
    \hline
    \multicolumn{7}{l}{\textbf{Phi-4}} \\
    \quad No mask & -- & 70.62 & 51.77 & 49.57 & 81.23 & 80.05 \\
    \quad Head mask, tuned on development data & 250 & 63.63 & 64.65 & 65.04 & 38.65 & 35.11 \\
    \quad Random heads, same layers (mean of 5) & 250 & 46.52 & 33.68 & 30.53 & 63.38 & 66.22 \\
    \cdashline{1-7}
    \quad No intervention & -- & 70.93 & 51.69 & 49.65 & 80.83 & 80.60 \\
    \quad \ours{} & -- & 69.05 & 66.77 & 65.20 & 72.51 & 71.48 \\
    \hline
    \multicolumn{7}{l}{\textbf{OLMo-2-7B-Instruct}} \\
    \quad No mask & -- & 42.97 & 23.72 & 17.91 & 70.07 & 76.83 \\
    \quad Head mask, tuned on development data & 137 & 40.53 & 40.69 & 41.08 & 27.26 & 24.90 \\
    \quad Random heads, same layers (mean of 5) & 137 & 28.47 & 21.84 & 19.15 & 42.44 & 48.28 \\
    \cdashline{1-7}
    \quad No intervention & -- & 42.42 & 23.33 & 16.42 & 70.62 & 78.63 \\
    \quad \ours{} & -- & 41.16 & 30.79 & 25.69 & 59.86 & 63.94 \\
    \hline
    \end{tabular}
    \caption{Head masks from the head search of \citet{liu2026sentence} and \ours{} (accuracy, \%). Tuned: chosen on development questions as described above; gold-cue control: the distractor points at the correct option. Random heads: as many heads removed from each layer as in the tuned mask, chosen at random (mean of five masks). Masked and unmasked rows are read in the same run. Mask rows use the forced-choice readout of the head search and \ours{} rows the first-token readout of Table~\ref{tab:main_acc}, so each is compared with the unmodified row of its own block.}
    \label{tab:headmask}
\end{table*}

\section{Analysis of \ours{}}
\label{app:sec_analysis}

\subsection{Batch Used for the Gate's Projection}
\label{app:basis}
\begin{table}[!htbp]
    \centering
    \small
    \setlength{\tabcolsep}{3pt}
    \resizebox{\columnwidth}{!}{
    \begin{tabular}{l c c c c}
    \hline
    & & \multicolumn{2}{c}{\textbf{\hard{}}} & \\
    \cmidrule(lr){3-4}
    \textbf{Batch for $P$} & \textbf{Clean} & \textbf{NL} & \textbf{BS} & \textbf{BMQ} \\
    \hline
    \multicolumn{5}{l}{\textit{Qwen3-8B}} \\
    \quad No intervention & 63.39 & 36.37 & 30.95 & 28.67 \\
    \quad Test set (\ours{}) & 61.74 & 58.68 & 54.20 & 53.50 \\
    \quad Batches of 50 questions & 61.78 & 58.76 & 53.74 & 53.31 \\
    \quad Fixed: dev.\ MedQA + \hard{} & 62.37 & 56.01 & 49.73 & 47.86 \\
    \hline
    \multicolumn{5}{l}{\textit{Llama-3.1-8B-Instruct}} \\
    \quad No intervention & 61.74 & 39.20 & 42.11 & 49.71 \\
    \quad Test set (\ours{}) & 59.62 & 58.21 & 56.56 & 55.41 \\
    \quad Batches of 50 questions & 59.96 & 58.17 & 56.45 & 55.44 \\
    \quad Fixed: dev.\ MedQA + \hard{} & 60.17 & 55.77 & 53.73 & 51.79 \\
    \hline
    \multicolumn{5}{l}{\textit{Llama-3-Meerkat-8B}} \\
    \quad No intervention & 63.00 & 37.47 & 38.02 & 42.69 \\
    \quad Test set (\ours{}) & 61.51 & 60.72 & 57.66 & 55.13 \\
    \quad Batches of 50 questions & 61.34 & 60.70 & 57.65 & 55.08 \\
    \quad Fixed: dev.\ MedQA + \hard{} & 60.88 & 59.54 & 55.46 & 52.27 \\
    \hline
    \multicolumn{5}{l}{\textit{Phi-4}} \\
    \quad No intervention & 70.93 & 51.69 & 49.65 & 57.05 \\
    \quad Test set (\ours{}) & 69.05 & 66.77 & 65.20 & 62.75 \\
    \quad Batches of 50 questions & 69.01 & 66.62 & 65.18 & 62.74 \\
    \quad Fixed: dev.\ MedQA + \hard{} & 68.66 & 63.63 & 63.55 & 60.69 \\
    \hline
    \end{tabular}%
    }
    \caption{Accuracy (\%) of \ours{} with the projection of the Discordance Gate estimated on different unlabeled batches. Batches of 50 questions: mean over 20 random partitions of each set. Fixed: estimated once on development (dev.) MedQA and \hard{} questions without labels and applied unchanged to every test set. BMQ: BiasMedQA, mean of seven types.}
    \label{tab:basis}
\end{table}
The projection $P$ of the Discordance Gate uses the mean and the first principal component of an unlabeled batch (Section~\ref{sec:method_dg}). The main results use all narrative sentences of each test set, without labels.
At the chosen layers, the first principal component carries 94--98\% of the variance of the sentence vectors for Llama-3.1-8B-Instruct, Qwen3-8B, Llama-3-Meerkat-8B and Phi-4, and 9\% for OLMo-2-7B-Instruct.
Table~\ref{tab:basis} keeps every hyperparameter and changes only this batch.
Random batches of 50 questions (about 400 sentences, 20 random partitions of each set) keep 98--100\% of the \hard{} gain, the worst partition staying within 0.8 points of the full set.
A projection fixed once on development clean and \hard{} questions and applied unchanged to every test set supports single-question inference and retains 79--95\% of the \hard{} accuracy gain obtained with the test-set projection.
Across the batch and fixed-projection variants, clean accuracy differs from the test-set projection by at most 0.7 points.

\subsection{Gate Representation}
\label{app:surface}
Table~\ref{tab:surface} replaces the hidden states behind the Discordance Gate by the embeddings of SimCSE \citep{gao-etal-2021-simcse}, MPNet \citep{song2020mpnet,reimers-gurevych-2019-sentence} or SapBERT \citep{liu-etal-2021-sapbert}: \hard{} accuracy falls by 4--15 points for both models, with clean accuracy within 1.2 points.

\begin{table*}[t]
    \centering
    \small
    \setlength{\tabcolsep}{4pt}
    \begin{tabular}{l c c c c c c c}
    \hline
    \multirow{2}{*}{\textbf{Variant}} & \multirow{2}{*}{\textbf{\#Fwd $\downarrow$}} & \multicolumn{2}{c}{\textbf{Clean}} & \multicolumn{2}{c}{\textbf{\hard{} NL}} & \multicolumn{2}{c}{\textbf{\hard{} BS}} \\
    \cmidrule(lr){3-4} \cmidrule(lr){5-6} \cmidrule(lr){7-8}
    & & Acc $\uparrow$ & Interv.\ rate $\downarrow$ & Acc $\uparrow$ & Interv.\ rate $\downarrow$ & Acc $\uparrow$ & Interv.\ rate $\downarrow$ \\
    \hline
    \multicolumn{8}{l}{\textbf{Replacing the representation of DG while IA stays fixed}} \\
    \multicolumn{8}{l}{\textit{Qwen3-8B}} \\
        Hidden states (Ours) & 25 & 61.74 & 19.40 & \textbf{58.68} & 47.84 & \textbf{54.20} & 49.33 \\
        SimCSE & 9$^\ddagger$ & 61.98 & 18.46 & 53.34 & 41.87 & 49.96 & 40.85 \\
        MPNet$^\dagger$ & 9$^\ddagger$ & \textbf{62.14} & \textbf{15.16} & 50.67 & 35.43 & 41.08 & \textbf{24.51} \\
        SapBERT$^\dagger$ & 9$^\ddagger$ & 62.06 & 16.03 & 43.28 & \textbf{24.82} & 42.18 & 28.52 \\
    \hline
    \multicolumn{8}{l}{\textit{Llama-3.1-8B-Instruct}} \\
        Hidden states (Ours) & 25 & 59.62 & 19.72 & \textbf{58.21} & 45.25 & \textbf{56.56} & 41.71 \\
        SimCSE & 9$^\ddagger$ & \textbf{60.80} & \textbf{11.31} & 49.65 & 28.04 & 50.27 & 25.14 \\
        MPNet$^\dagger$ & 9$^\ddagger$ & 59.94 & 14.93 & 51.14 & 33.54 & 47.68 & \textbf{23.64} \\
        SapBERT$^\dagger$ & 9$^\ddagger$ & 59.70 & 17.36 & 44.85 & \textbf{24.74} & 47.53 & 25.37 \\
    \hline
    \end{tabular}
    \caption{The Discordance Gate computed from an external sentence encoder's embeddings instead of the model's hidden states, with Interventional Attribution and the blocking unchanged (columns as in Table~\ref{tab:ablation}): SimCSE \citep{gao-etal-2021-simcse} with its threshold chosen on the development set, MPNet (all-mpnet-base-v2, a Sentence-BERT model; \citealp{song2020mpnet,reimers-gurevych-2019-sentence}) and SapBERT \citep{liu-etal-2021-sapbert}; $^\dagger$threshold not tuned (that of \ours{}); $^\ddagger$model forward passes for IA alone, plus two passes of the encoder per sentence.}
    \label{tab:surface}
\end{table*}

\subsection{How the Gate Sees the Distractor}
\label{app:regime}
\begin{table*}[!htbp]
    \centering
    \small
    \setlength{\tabcolsep}{3pt}
    \resizebox{\linewidth}{!}{
    \begin{tabular}{l r r r r r}
    \hline
    \multicolumn{6}{l}{\textit{(a) How the gate sees the distractor (\% of flipped questions).}} \\
    \textbf{Set} & \textbf{Attacked} & \textbf{Too short} & \textbf{(i) Flagged first} & \textbf{(ii) Flagged, not first} & \textbf{(iii) Not flagged} \\
    \hline
    \multicolumn{6}{l}{\textbf{Qwen3-8B}} \\
    \hard{} NL & 353 & 2.5 & 87.3 & 4.8 & 5.4 \\
    \hard{} BS & 421 & 1.9 & 69.1 & 11.9 & 17.1 \\
    MedDistractQA NL & 209 & 2.4 & 90.9 & 4.3 & 2.4 \\
    MedDistractQA BS & 202 & 2.0 & 83.7 & 5.4 & 8.9 \\
    BiasMedQA (7 types) & 2,661 & 1.5 & 85.3 & 6.3 & 6.8 \\
    \multicolumn{6}{l}{\textbf{Llama-3.1-8B-Instruct}} \\
    \hard{} NL & 312 & 1.0 & 91.7 & 2.9 & 4.5 \\
    \hard{} BS & 286 & 1.4 & 75.2 & 10.8 & 12.6 \\
    MedDistractQA NL & 155 & 0.0 & 89.0 & 3.9 & 7.1 \\
    MedDistractQA BS & 131 & 0.0 & 92.4 & 0.8 & 6.9 \\
    BiasMedQA (7 types) & 1,044 & 0.8 & 92.3 & 3.2 & 3.7 \\
    \hline
    \multicolumn{6}{l}{\textit{(b) Questions recovered (\%), \hard{}, MedDistractQA and BiasMedQA pooled}} \\
    \textbf{Case} & & \textbf{\ours{} w/o IA} & \textbf{\ours{} w/o DG} & \textbf{\ours{}} & \textbf{Known distractor} \\
    \hline
    \multicolumn{6}{l}{\textbf{Qwen3-8B}} \\
    (i) Flagged first & (3,229) & 93 & 91 & 88 & 93 \\
    (ii) Flagged, not first & (254) & 3 & 94 & 93 & 96 \\
    (iii) Not flagged & (296) & 8 & 87 & 2 & 88 \\
    \multicolumn{6}{l}{\textbf{Llama-3.1-8B-Instruct}} \\
    (i) Flagged first & (1,724) & 89 & 84 & 73 & 89 \\
    (ii) Flagged, not first & (80) & 8 & 89 & 79 & 91 \\
    (iii) Not flagged & (109) & 6 & 87 & 3 & 96 \\
    \hline
    \end{tabular}
    }
    \caption{How the gate sees the distractor on the flipped questions, answered correctly without the distractor and wrongly with it, with the distractor located. Too short: fewer than three narrative sentences, so the gate is not applied. (i) Flagged first: the distractor passes the gate and is the most discordant sentence. (ii) Flagged, not first. (iii) Not flagged.}
    \label{tab:regime}
\end{table*}
Interventional Attribution resolves a specific ambiguity that discordance alone cannot (Table~\ref{tab:regime}; Section~\ref{sec:selectivity}). Among the flipped questions pooled across \hard{}, MedDistractQA, and BiasMedQA, the distractor passes the gate but is not the most discordant sentence in 254 of 3,846 Qwen questions and 80 of 1,928 Llama questions. In this subset, attribution ranks the distractor first in 245/254 and 66/80 questions, respectively. Blocking the most discordant sentence recovers only 3\% and 8\%, whereas \ours{} recovers 93\% and 79\%. This benefit has a cost: when the distractor is already the most discordant sentence, \ours{} w/o IA blocking recovers 93\% and 89\%, compared with 88\% and 73\% for \ours{}. Attribution therefore adds source localization in the ambiguous subset, rather than a uniform accuracy gain over the \ours{} w/o IA variant.

\subsection{Added Off-Case Sentences}
\label{app:decoy}
For each \hard{} test question, we insert the distractor of another question of the same type (Nonliteral or Bystander), chosen at random among those that share no content word with any of the question's options.
Content words are words of at least four letters other than stop words, and two words count as shared when their first five letters match, following the five-letter truncation of \citet{mcnamee2009addressing}, so that, for example, ``fracture'' and ``fractured'' match.
The added sentence is inserted at a random position after the first sentence and before the question's own distractor, which stays where it is in the main experiments.
The sets with two and three added sentences each add one further sentence to the previous set, chosen and inserted in the same way.
We exclude 47 Nonliteral questions whose options contain ``patient'', which appears in nearly all Nonliteral distractors and leaves no eligible off-case remark that shares no content word with the options.
We also exclude 18 Nonliteral and 19 Bystander questions with at most one sentence before the original distractor, leaving no valid insertion position.
This leaves 1,208 Nonliteral and 1,254 Bystander questions in every set.
With one added sentence, blocking the added sentence alone changes the answer about as often as blocking an ordinary sentence of the case (10--14\% against 8--10\% of questions) and far less often than blocking the distractor (31--42\%), indicating that the added sentence has a much smaller effect on the model's decision.
All models use the hyperparameters of Table~\ref{tab:hparams}, and the gate's projection is estimated from each set's own sentences as in the main experiments.

Table~\ref{tab:decoy} gives the results.
Without intervention, accuracy rises by 3.6--6.9 points from zero to three added sentences, as the added sentences weaken the distractor's influence: the share of answers selecting its target option falls by 5.7--9.4 points, and the distractor's median attribution score falls by half or more. 
In the same setting, the attention that the answer position pays to the distractor, averaged over layers and heads, falls by 22--32\% from zero to three added sentences for every model and distractor type.
Across questions, a larger fall from zero to one added sentence is correlated with a larger fall in the target-minus-gold logit gap (Spearman's $\rho$ 0.18--0.37 \citep{spearman1904proof}, with lower bounds of the 95\% bootstrap intervals of 0.12--0.32).

For each noise level, flipped questions are those answered correctly on the original clean question but incorrectly on the corresponding augmented question.
The denominator is recomputed at each noise level and includes questions without intervention.
With one, two and three added sentences, \ours{} blocks the original distractor in 54--80\%, 41--67\% and 35--64\% of flipped questions.
The corresponding rates for \ours{} w/o IA are 24--31\%, 16--22\% and 12--18\%, while it blocks an added sentence instead in 59--67\%, 70--80\% and 77--86\%.

\begin{table*}[t]
    \centering
    \begin{tabular}{l l c c c c c c}
    \hline
    \textbf{Model} & & \textbf{+} & \textbf{None} & \textbf{w/o IA} & \textbf{\ours{}} & \textbf{Known} & \textbf{\ours{} $-$ w/o IA} \\
    \hline
    Llama-3.1-8B-Instruct & NL & 1 & 42.7 & 45.2 & 53.3 & 60.2 & +8.1 [6.1, 10.2] \\
     & NL & 2 & 44.2 & 46.9 & 52.7 & 60.8 & +5.9 [4.0, 7.8] \\
     & NL & 3 & 45.0 & 46.3 & 53.3 & 59.9 & +7.0 [5.1, 9.0] \\
     & BS & 1 & 45.4 & 45.7 & 51.9 & 60.1 & +6.2 [4.4, 8.1] \\
     & BS & 2 & 45.3 & 46.3 & 51.8 & 58.5 & +5.6 [3.8, 7.3] \\
     & BS & 3 & 45.5 & 47.0 & 51.0 & 59.0 & +4.0 [2.4, 5.7] \\
    \hline
    Qwen3-8B & NL & 1 & 38.3 & 43.6 & 52.0 & 60.2 & +8.4 [6.4, 10.4] \\
     & NL & 2 & 41.5 & 42.7 & 51.2 & 60.3 & +8.4 [6.6, 10.4] \\
     & NL & 3 & 41.9 & 43.2 & 50.4 & 60.1 & +7.2 [5.4, 9.1] \\
     & BS & 1 & 34.2 & 38.7 & 48.9 & 61.3 & +10.2 [8.1, 12.4] \\
     & BS & 2 & 36.1 & 38.7 & 47.5 & 61.5 & +8.9 [6.8, 11.0] \\
     & BS & 3 & 36.6 & 39.2 & 45.4 & 60.1 & +6.2 [4.5, 8.1] \\
    \hline
    Llama-3-Meerkat-8B & NL & 1 & 42.8 & 43.9 & 54.4 & 61.6 & +10.5 [8.3, 12.8] \\
     & NL & 2 & 42.3 & 45.5 & 54.0 & 60.4 & +8.5 [6.3, 10.8] \\
     & NL & 3 & 43.0 & 45.8 & 55.5 & 60.4 & +9.7 [7.5, 11.8] \\
     & BS & 1 & 40.4 & 44.3 & 53.8 & 61.5 & +9.5 [7.3, 11.7] \\
     & BS & 2 & 41.6 & 44.6 & 54.1 & 61.2 & +9.5 [7.4, 11.6] \\
     & BS & 3 & 43.1 & 45.0 & 54.2 & 60.0 & +9.3 [7.1, 11.4] \\
    \hline
    \end{tabular}
    \caption{Accuracy (\%) on \hard{} with added off-case sentences. +: number of added sentences. None: no intervention. w/o IA: \ours{} w/o IA. Known: the question's own distractor blocked. Last column: difference in points with paired 95\% bootstrap interval.}
    \label{tab:decoy}
\end{table*}

\subsection{When \ours{} Intervenes}
\label{app:when}
\ours{} responds to the distractor's effect on the answer, not to its presence (Table~\ref{tab:when}; Section~\ref{sec:selectivity}).
It intervenes on 79--94\% of the questions whose answer the distractor pulled to the target, on 25--38\% of the other wrong answers and on 8--22\% of the correct ones.
On questions answered correctly both with and without the distractor, it intervenes less often with the distractor (7--21\%) than on their clean versions (19--30\%).
The answers pulled to the target differ between the two models (Jaccard 0.35--0.36), and \ours{} w/o IA intervenes on 92--99\% of all questions (Table~\ref{tab:ablation}), so its intervention carries no such information.

\begin{table*}[t]
    \centering
    \small
    \setlength{\tabcolsep}{3.5pt}
    \begin{tabular}{l cc cc cc}
    \hline
    & \multicolumn{2}{c}{\textbf{Llama-3.1-8B-Instruct}} & \multicolumn{2}{c}{\textbf{Qwen3-8B}} & \multicolumn{2}{c}{\textbf{Llama-3-Meerkat-8B}} \\
    \cmidrule(lr){2-3} \cmidrule(lr){4-5} \cmidrule(lr){6-7}
    \textbf{Answer without intervention} & \textbf{NL} & \textbf{BS} & \textbf{NL} & \textbf{BS} & \textbf{NL} & \textbf{BS} \\
    \hline
    \multicolumn{7}{l}{\textit{\ours{} intervenes on (\%, questions in the group)}} \\
    Target, pulled by the distractor & 88.5 (390) & 81.0 (343) & 88.5 (443) & 79.0 (538) & 94.0 (450) & 88.6 (437) \\
    Target, already on the clean question & 67.4 (193) & 53.3 (199) & 63.4 (232) & 57.5 (219) & 71.5 (200) & 68.3 (189) \\
    Wrong, not the target & 27.7 (191) & 33.3 (195) & 25.2 (135) & 26.2 (122) & 31.5 (146) & 38.0 (163) \\
    Correct & 9.6 (499) & 15.3 (536) & 7.8 (463) & 11.4 (394) & 20.1 (477) & 21.7 (484) \\
    \cdashline{1-7}
    \multicolumn{7}{l}{\textit{Same questions, answered correctly with and without the distractor}} \\
    \quad with the distractor & 7.8 (472) & 12.0 (500) & 6.9 (452) & 10.9 (384) & 18.1 (441) & 20.5 (458) \\
    \quad without it (clean version) & 19.3 & 19.6 & 20.1 & 20.1 & 30.2 & 28.6 \\
    \hline
    \multicolumn{7}{l}{\textit{Target-wrong answers (\%)}} \\
    \ours{} intervenes & 82.5 & 72.3 & 88.5 & 87.7 & 79.9 & 75.6 \\
    \ours{} does not intervene & 15.5 & 21.3 & 20.5 & 31.9 & 14.9 & 18.6 \\
    All questions & 45.8 & 42.6 & 53.0 & 59.5 & 51.1 & 49.2 \\
    \hline
    \end{tabular}
    \caption{How often \ours{} intervenes (\%; questions in parentheses), by the answer each model gives on \hard{} without intervention. Pulled by the distractor: the target on \hard{} but not on the clean question; already on the clean question: the target even without the distractor. Groups use the target labels, which \ours{} never sees. Lower block: target-wrong answers among the questions on which \ours{} intervenes or not.}
    \label{tab:when}
\end{table*}

Table~\ref{tab:funnel_full} traces the successive steps of \ours{} on flipped questions from \hard{} and MedDistractQA.
\begin{table}[!htbp]
    \centering
    \small
    \setlength{\tabcolsep}{3pt}
    \resizebox{\columnwidth}{!}{%
    \begin{tabular}{l c c c c c}
    \hline
    & \textbf{IA ranks} & \textbf{DG} & \textbf{Blocking} & \multicolumn{2}{c}{\textbf{Recovered}} \\
    \cmidrule(lr){5-6}
    \textbf{Set} & \textbf{it first} & \textbf{passes it} & \textbf{restores} & \textbf{\ours{}} & \textbf{Known} \\
    \hline
    \multicolumn{6}{l}{\textit{Llama-3.1-8B-Instruct}} \\
    \hard{} NL & 90 & 95 & 95 & 81 & 94 \\
    \hard{} BS & 90 & 86 & 93 & 73 & 92 \\
    MedDistractQA NL & 72 & 93 & 90 & 62 & 88 \\
    MedDistractQA BS & 76 & 97 & 90 & 67 & 85 \\
    \multicolumn{6}{l}{\textit{Qwen3-8B}} \\
    \hard{} NL & 91 & 92 & 93 & 78 & 90 \\
    \hard{} BS & 90 & 81 & 95 & 71 & 93 \\
    MedDistractQA NL & 89 & 95 & 93 & 79 & 89 \\
    MedDistractQA BS & 83 & 90 & 88 & 67 & 85 \\
    \hline
    \end{tabular}%
    }
    \caption{Steps of \ours{} on flipped questions from \hard{} and MedDistractQA (\%). The first three columns give successive pass rates. Recovered: the share of all flipped questions answered correctly after \ours{} or blocking the known distractor.}
    \label{tab:funnel_full}
\end{table}

\subsection{Clean Cost}
\label{app:cleancost}
The rate of damage is similar across clean sets: on Llama, \ours{} repairs 4.7\% of the wrongly answered and damages 6.4\% of the correctly answered MedQA questions, and 4.6\% and 6.8\% on MMLU professional medicine, so the net loss grows with baseline accuracy (2.1 points on MedQA at 61.7\%, 4.0 points on MMLU-PM at 76.1\%).

\section{Additional Results}
\label{app:sec_results}

\subsection{Performance by Distractor Type}
\label{app:results_by_type}
Table~\ref{tab:main_acc_full} reports the Nonliteral and Bystander results underlying the averages in Table~\ref{tab:main_acc}.
\begin{table*}[!t]
    \centering
    \small
    \setlength{\tabcolsep}{3.3pt}
    \begin{tabular}{l c c c c c c c c c}
    \hline
    \multirow{2}{*}{\textbf{Method}} & \multirow{2}{*}{\textbf{\#Fwd}} & \multirow{2}{*}{\textbf{s/q}}
    & \multicolumn{2}{c}{\textbf{Clean}} & \multicolumn{2}{c}{\textbf{\hard}} & \multicolumn{2}{c}{\textbf{MedDistractQA}} & \textbf{BiasMedQA} \\
    \cmidrule(lr){4-5} \cmidrule(lr){6-7} \cmidrule(lr){8-9} \cmidrule(lr){10-10}
    & & & \textbf{MedQA} & \textbf{MMLU-PM} & \textbf{NL} & \textbf{BS} & \textbf{NL} & \textbf{BS} & \textbf{Avg.} \\
    \hline
    \multicolumn{10}{l}{\textit{Qwen3-8B}} \\
        No intervention & 1 & 0.041
        & 63.39 & 82.72 & 36.37 & 30.95 & 49.41 & 51.14 & 28.67 \\
        Ignore instruction & 1 & 0.044
        & \textbf{64.02} & 82.72 & 35.35 & 31.03 & 49.49 & 51.22 & 27.08 \\
        CAA & 1 & 0.041
        & \textbf{64.02} & 80.51 & 40.46 & 36.21 & 51.77 & 54.83 & 35.57 \\
        S2A$^\dagger$ & -- & 4.182
        & 63.63 & \textbf{83.46} & 39.75 & 33.23 & 55.93 & 52.63 & 35.34 \\
        \ours{} (Ours) & 25 & 0.819
        & 61.74 & 83.09 & \textbf{58.68} & \textbf{54.20} & \textbf{61.74} & \textbf{60.17} & \textbf{53.50} \\
    \cdashline{1-10}
        CoT & -- & 3.408
        & \textbf{70.23} & \textbf{86.03} & 46.27 & 42.11 & 57.19 & 59.15 & 41.51 \\
        CoT + I$^3$C & -- & 11.322
        & 65.59 & 84.93 & 39.83 & 39.12 & 53.18 & 57.50 & 35.01 \\
        CoT + \ours{} (Ours) & -- & 4.348
        & 67.79 & 85.66 & \textbf{63.00} & \textbf{59.94} & \textbf{64.02} & \textbf{66.30} & \textbf{57.98} \\
    \hline
    \multicolumn{10}{l}{\textit{Llama-3.1-8B-Instruct}} \\
        No intervention & 1 & 0.040
        & 61.74 & \textbf{76.10} & 39.20 & 42.11 & 53.10 & 55.93 & 49.71 \\
        Ignore instruction & 1 & 0.041
        & 61.90 & \textbf{76.10} & 39.36 & 41.56 & 54.05 & 55.93 & 49.19 \\
        CAA & 1 & 0.040
        & \textbf{62.29} & 75.00 & 40.61 & 42.89 & 54.28 & 56.48 & 50.69 \\
        S2A$^\dagger$ & -- & 3.173
        & 61.12 & 75.74 & \textbf{58.29} & 53.10 & \textbf{60.25} & \textbf{60.02} & \textbf{56.83} \\
        \ours{} (Ours) & 25 & 0.731
        & 59.62 & 72.06 & 58.21 & \textbf{56.56} & 59.07 & \textbf{60.02} & 55.41 \\
    \cdashline{1-10}
        CoT & -- & 2.049
        & \textbf{67.71} & \textbf{79.04} & 45.40 & 50.04 & 61.19 & 62.06 & 53.30 \\
        CoT + I$^3$C & -- & 10.061
        & 67.01 & 78.31 & 46.58 & 47.45 & 55.93 & 59.78 & 47.20 \\
        CoT + \ours{} (Ours) & -- & 2.857
        & 66.46 & 75.37 & \textbf{61.51} & \textbf{60.88} & \textbf{65.44} & \textbf{64.89} & \textbf{57.56} \\
    \hline
    \multicolumn{10}{l}{\textit{Llama-3-Meerkat-8B}} \\
        No intervention & 1 & 0.038
        & 63.00 & 77.94 & 37.47 & 38.02 & 52.08 & 55.77 & 42.69 \\
        Ignore instruction & 1 & 0.040
        & 63.32 & 77.21 & 37.47 & 37.71 & 50.27 & 55.30 & 41.76 \\
        CAA & 1 & 0.038
        & 63.63 & 75.37 & 38.49 & 40.85 & 53.10 & 56.64 & 45.82 \\
        S2A$^\dagger$ & -- & 3.324
        & \textbf{63.94} & \textbf{79.04} & 46.19 & 45.40 & 57.42 & 58.99 & 48.55 \\
        \ours{} (Ours) & 25 & 0.721
        & 61.51 & 75.37 & \textbf{60.72} & \textbf{57.66} & \textbf{62.22} & \textbf{60.64} & \textbf{55.13} \\
    \cdashline{1-10}
        CoT$^\ddagger$ & -- & 6.078
        & \textbf{72.74} & \textbf{83.46} & 46.90 & 50.04 & 61.12 & 65.83 & 58.56 \\
        CoT + I$^3$C$^\ddagger$ & -- & 20.269
        & 64.73 & 82.35 & 42.58 & 43.44 & 55.46 & 57.66 & 45.43 \\
        CoT + \ours{} (Ours)$^\ddagger$ & -- & 7.301
        & 71.01 & 78.68 & \textbf{65.51} & \textbf{63.32} & \textbf{67.32} & \textbf{67.64} & \textbf{64.36} \\
    \hline
    \end{tabular}
    \caption{Accuracy (\%) by distractor type for the methods in Table~\ref{tab:main_acc}.  MMLU-PM: MMLU professional medicine. BiasMedQA: mean of seven bias types. NL / BS: Nonliteral / Bystander. \#Fwd: forward passes per question, not counting the generation of S2A$^\dagger$ \citep{weston2023s2a}. s/q: seconds per question, including generation. Below the dashed line: chain-of-thought generation (Appendix~\ref{app:generation}), with I$^3$C \citep{wu2024instructing}. $^\ddagger$Prompted in the answer format of its model card (Appendix~\ref{app:prompts}). \textbf{Bold}: best per model and column, above and below the dashed line separately.}
    \label{tab:main_acc_full}
\end{table*}

\subsection{Context Attribution without the Gate}
\label{app:top1}
\begin{table*}[!t]
    \centering
    \small
    \setlength{\tabcolsep}{4pt}
    \begin{tabular}{l c c c c c c c}
    \hline
    \multirow{2}{*}{\textbf{Variant}} & \multirow{2}{*}{\textbf{\#Fwd $\downarrow$}} & \multicolumn{2}{c}{\textbf{Clean}} & \multicolumn{2}{c}{\textbf{\hard{} NL}} & \multicolumn{2}{c}{\textbf{\hard{} BS}} \\
    \cmidrule(lr){3-4} \cmidrule(lr){5-6} \cmidrule(lr){7-8}
    & & Acc $\uparrow$ & Interv.\ rate $\downarrow$ & Acc $\uparrow$ & Interv.\ rate $\downarrow$ & Acc $\uparrow$ & Interv.\ rate $\downarrow$ \\
    \hline
    \multicolumn{8}{l}{\textit{Qwen3-8B}} \\
        IA (Ours) & 25 & 61.74 & \textbf{19.40} & \textbf{58.68} & 47.84 & 54.20 & 49.33 \\
        Avg.\ Attention & 18 & \textbf{62.84} & 35.74 & 37.63 & \textbf{14.53} & 31.03 & \textbf{20.03} \\
        AT2 & -- & -- & -- & -- & -- & -- & -- \\
        CAMAB & 58 & 62.22 & 19.95 & 54.75 & 39.59 & 51.45 & 43.05 \\
        ContextCite & 82 & 62.29 & \textbf{19.48} & 58.21 & 49.02 & \textbf{54.83} & 51.06 \\
    \hline
    \multicolumn{8}{l}{\textit{Llama-3.1-8B-Instruct}} \\
        IA (Ours) & 25 & 59.62 & 19.72 & \textbf{58.21} & 45.25 & \textbf{56.56} & 41.71 \\
        Avg.\ Attention & 18 & \textbf{61.04} & 20.97 & 55.93 & 69.21 & 51.85 & 42.81 \\
        AT2 & 18 & 60.72 & 20.42 & 53.97 & 55.62 & 48.47 & \textbf{29.46} \\
        CAMAB & 58 & 60.25 & 19.32 & 55.46 & \textbf{39.12} & 54.52 & 35.66 \\
        ContextCite & 82 & 59.62 & \textbf{19.09} & 57.89 & 43.75 & 55.70 & 39.51 \\
    \hline
    \end{tabular}
    \caption{Replacing IA while keeping the Discordance Gate fixed. Acc: accuracy (\%); Interv.: questions intervened on (\%); \#Fwd as in Table~\ref{tab:main_acc}. AT2 is evaluated on Llama-3.1-8B-Instruct only (--).}
    \label{tab:selector_gate}
\end{table*}

Each attribution method below blocks its top-ranked sentence in every question, with the same blocking as \ours{} and columns as in Table~\ref{tab:main_acc}; Table~\ref{tab:selector_gate} places the same methods under the Discordance Gate instead of Interventional Attribution.
These methods lose 3--10 points on clean questions. On \hard{}, averaged attention comes closest for Llama and Meerkat (3.5--7.7 points behind \ours{}), an advantage tied to the distractor's place at the end of the narrative (Appendix~\ref{app:position}). CAMAB and ContextCite remain 6--12 points behind.
\begin{table*}[t]
    \centering
    \small
    \setlength{\tabcolsep}{4pt}
    \begin{tabular}{l c c c c c c c c c}
    \hline
    \multirow{2}{*}{\textbf{Method}} & \multirow{2}{*}{\textbf{\#Fwd}} & \multirow{2}{*}{\textbf{s/q}}
    & \multicolumn{2}{c}{\textbf{Clean}} & \multicolumn{2}{c}{\textbf{\hard}} & \multicolumn{2}{c}{\textbf{MedDistractQA}} & \textbf{BiasMedQA} \\
    \cmidrule(lr){4-5} \cmidrule(lr){6-7} \cmidrule(lr){8-9} \cmidrule(lr){10-10}
    & & & \textbf{MedQA} & \textbf{MMLU-PM} & \textbf{NL} & \textbf{BS} & \textbf{NL} & \textbf{BS} & \textbf{Avg.} \\
    \hline
    \multicolumn{10}{l}{\textit{Qwen3-8B}} \\
        Avg.\ Attention & 2 & 0.067
        & 60.25 & 81.99 & 33.86 & 26.71 & 46.11 & 46.98 & 45.89 \\
        CAMAB & 42 & 1.39
        & 55.93 & 73.90 & 46.35 & 47.29 & 49.73 & 49.49 & 43.33 \\
        ContextCite & 66 & 1.84
        & 53.57 & 71.69 & 47.45 & 48.23 & 48.63 & 47.53 & 43.77 \\
    \hline
    \multicolumn{10}{l}{\textit{Llama-3.1-8B-Instruct}} \\
        Avg.\ Attention & 2 & 0.059
        & 56.17 & 72.43 & 54.44 & 48.86 & 55.85 & 54.99 & {55.84} \\
        AT2 & 2 & 0.069
        & 56.95 & 72.43 & 51.14 & 45.40 & 54.99 & 54.12 & 48.18 \\
        CAMAB & 42 & 1.21
        & 52.87 & 70.22 & 47.92 & 47.53 & 51.37 & 51.53 & 44.01 \\
        ContextCite & 66 & 1.75
        & 51.77 & 67.65 & 48.15 & 48.55 & 49.96 & 50.67 & 43.44 \\
    \hline
    \multicolumn{10}{l}{\textit{Llama-3-Meerkat-8B}} \\
        Avg.\ Attention & 2 & --
        & 59.47 & 72.79 & 57.27 & 52.08 & 58.60 & 55.62 & {56.65} \\
        CAMAB & 42 & --
        & 56.01 & 70.59 & 51.22 & 49.65 & 52.40 & 52.47 & 44.55 \\
        ContextCite & 66 & --
        & 54.75 & 66.54 & 51.14 & 50.67 & 51.69 & 50.35 & 43.36 \\
    \hline
    \end{tabular}
    \caption{Context attribution methods that block their top-ranked sentence in every question (accuracy, \%; columns as in Table~\ref{tab:main_acc}, where \ours{} and the other baselines are reported).}
    \label{tab:top1}
\end{table*}

\subsection{Generation-Based Prompting}
\label{app:generation}
\begin{figure*}[t]
    \centering
    \includegraphics[width=\linewidth]{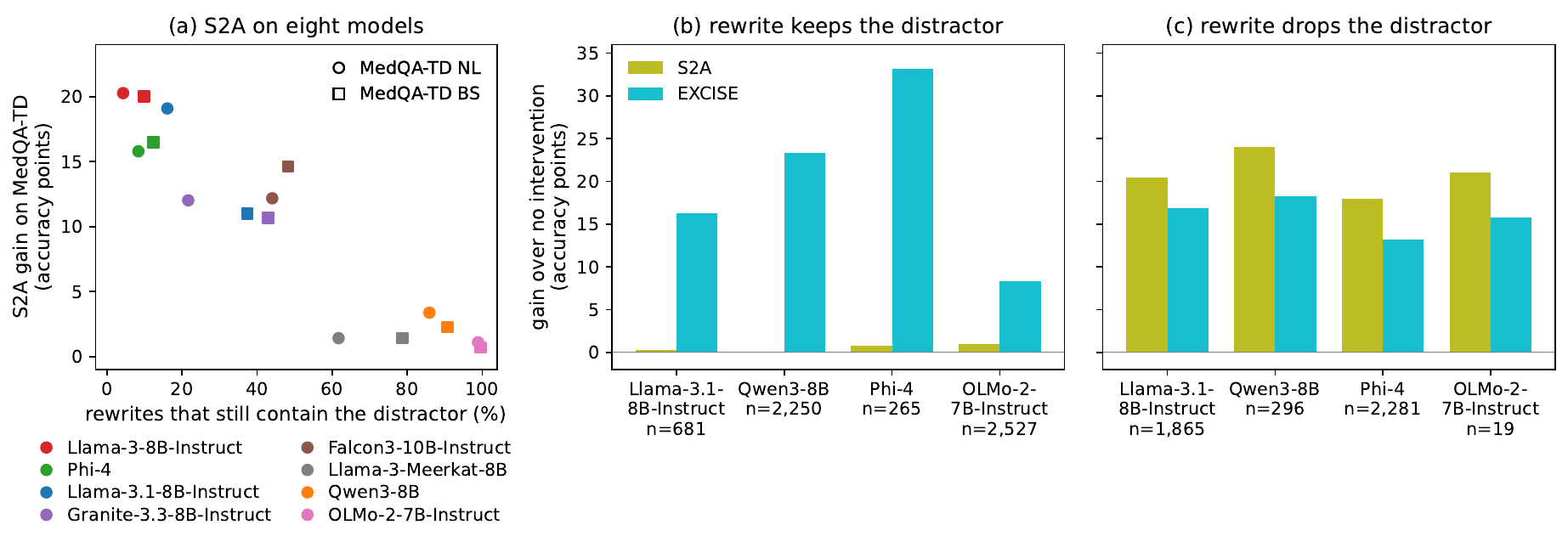}
    \caption{Gains of S2A \citep{weston2023s2a} by whether the model's rewrite keeps the distractor (\hard{}). (a) Share of S2A rewrites that keep the distractor (at least 80\% of its content words) against the gain of S2A over no intervention, for eight instruct models and both distractor types. (b, c) Gains of S2A and \ours{} on the questions whose rewrite keeps (b) or drops (c) the distractor, for the four models run with both methods (Nonliteral and Bystander pooled; $n$: questions).}
    \label{fig:s2a_filtering}
\end{figure*}

S2A answers from its rewrite with the same first-token readout, and Table~\ref{tab:s2a_excise} gives its accuracy with the share of rewrites that keep the distractor. Figure~\ref{fig:s2a_filtering} relates the gain of S2A to whether the rewrite keeps the distractor.
For Llama-3.1-8B-Instruct and Qwen3-8B, chain-of-thought generation uses the system and user messages of the answer readout (Appendix~\ref{app:prompts}) without the fixed assistant prefix, and the answer is read from the generated JSON object's ``answer'' field.
All chain-of-thought outputs are decoded greedily with at most 1,024 new tokens.
For BiasMedQA, whose questions have five options, the instruction lists the letters A to E. The MMLU-PM and MedDistractQA columns of Qwen3-8B and Llama-3.1-8B-Instruct and all of Llama-3-Meerkat-8B's are decoded with the key-value cache, which gives the same answer as decoding without it for 92--93\% of 100 \hard{} questions for Qwen3-8B and Llama-3.1-8B-Instruct (accuracy within 2 points) and for all of 10 questions for Llama-3-Meerkat-8B. Llama-3-Meerkat-8B ignores the shared JSON answer format, so it is prompted with the system message of its model card (Appendix~\ref{app:prompts}) and its answer is read from the last ``the answer is (X)'', which is also accepted without the opening parenthesis. The same rule reads every Llama-3-Meerkat-8B row. It reads 93--98\% of the CoT + I$^3$C answers and at least 98.9\% of the other answers. Requiring the parenthesis would lower CoT + I$^3$C by 2.7--5.1 points and leave the other rows unchanged. Its I$^3$C verifications follow the requested format in 37\% of cases (Llama-3.1-8B-Instruct: 99\%), so its CoT + I$^3$C row shows I$^3$C applied to a model that rarely completes the verification step.
Applied to the model's own S2A rewrite, \ours{} stays within 1.7 points of the better single method in every \hard{} cell of Llama-3.1-8B-Instruct, Qwen3-8B and Phi-4 and inherits the clean cost of \ours{} (Table~\ref{tab:s2a_excise}).

\begin{table*}[t]
    \centering
    \small
    \setlength{\tabcolsep}{3.5pt}
    \resizebox{\linewidth}{!}{%
    \begin{tabular}{l ccccc ccccc cccc}
    \hline
    & \multicolumn{5}{c}{\textbf{\hard{} NL}} & \multicolumn{5}{c}{\textbf{\hard{} BS}} & \multicolumn{4}{c}{\textbf{Clean MedQA}} \\
    \cmidrule(lr){2-6} \cmidrule(lr){7-11} \cmidrule(lr){12-15}
    \textbf{Model} & Kept & None & S2A & \ours{} & S2A+\ours{} & Kept & None & S2A & \ours{} & S2A+\ours{} & None & S2A & \ours{} & S2A+\ours{} \\
    \hline
    Llama-3.1-8B-Instruct & 16.1 & 39.2 & \textbf{58.3} & 58.2 & 57.8 & 37.4 & 42.1 & 53.1 & \textbf{56.6} & 54.9 & \textbf{61.7} & 61.1 & 59.6 & 59.3 \\
    Qwen3-8B & 85.9 & 36.4 & 39.7 & 58.7 & \textbf{59.3} & 90.8 & 31.0 & 33.2 & 54.2 & \textbf{54.8} & 63.4 & \textbf{63.6} & 61.7 & 62.1 \\
    Phi-4 & 8.4 & 51.7 & \textbf{67.5} & 66.8 & 66.5 & 12.4 & 49.6 & \textbf{66.1} & 65.2 & 65.5 & 70.9 & \textbf{71.5} & 69.1 & 68.7 \\
    OLMo-2-7B-Instruct & 98.9 & 23.3 & 24.3 & \textbf{30.8} & 26.6 & 99.6 & 16.4 & 17.7 & \textbf{25.7} & 20.4 & \textbf{42.4} & -- & 41.2 & -- \\
    \hline
    \end{tabular}%
    }
    \caption{S2A \citep{weston2023s2a}, \ours{}, and \ours{} applied to the model's own S2A rewrite (S2A+\ours{}; an exploratory combination). Accuracy (\%). Kept: share (\%) of S2A rewrites that keep the distractor, i.e., contain at least 80\% of its content words. \textbf{Bold}: highest accuracy per model and set; --: S2A rewrites not produced for OLMo-2-7B-Instruct's clean questions.}
    \label{tab:s2a_excise}
\end{table*}

\subsection{BiasMedQA by Bias Type}
\label{app:biasmedqa}
\ours{} raises BiasMedQA accuracy on all seven bias types for Qwen3-8B (+14.8 to +31.3 points) and on six of seven for Llama-3.1-8B-Instruct (+1.9 to +14.2) and Llama-3-Meerkat-8B (+6.0 to +32.9), and leaves confirmation unchanged within its interval for both (Table~\ref{tab:biasmedqa}).
The gate flags 89--97\% of the bias cues, so the remaining gap comes from attribution, which ranks the cue first in 19--42\% of questions for Llama and 39--74\% for Qwen.
For Llama-3-Meerkat-8B, the mean over the seven types rises from 42.7 to 55.1 and confirmation changes by $-$1.0 [$-$2.7, +0.7].
\begin{table*}[t]
    \centering
    \small
    \setlength{\tabcolsep}{4pt}
    \resizebox{\linewidth}{!}{%
    \begin{tabular}{l ccccc ccccc}
    \hline
    & \multicolumn{5}{c}{\textbf{Llama-3.1-8B-Instruct}} & \multicolumn{5}{c}{\textbf{Qwen3-8B}} \\
    \cmidrule(lr){2-6} \cmidrule(lr){7-11}
    \textbf{Bias type} & No int. & \ours{} & $\Delta$ [95\% CI] & IA top-1 & DG flag & No int. & \ours{} & $\Delta$ [95\% CI] & IA top-1 & DG flag \\
    \hline
    Recency & 49.6 & 54.3 & +4.7 [+3.1, +6.4] & 26 & 95 & 25.9 & 52.9 & +27.0 [+24.5, +29.6] & 63 & 93 \\
    Confirmation & 59.1 & 57.7 & $-$1.3 [$-$2.7, +0.1] & 19 & 96 & 41.1 & 55.9 & +14.8 [+12.5, +17.1] & 39 & 94 \\
    Frequency & 38.2 & 52.4 & +14.2 [+12.0, +16.4] & 42 & 95 & 25.6 & 52.9 & +27.3 [+24.7, +29.9] & 65 & 89 \\
    Status quo & 44.4 & 54.1 & +9.7 [+7.7, +11.9] & 36 & 94 & 27.7 & 53.1 & +25.5 [+22.9, +28.1] & 63 & 93 \\
    False consensus & 52.0 & 56.8 & +4.8 [+3.0, +6.6] & 26 & 97 & 20.3 & 51.4 & +31.0 [+28.3, +33.7] & 74 & 90 \\
    Self-diagnosis & 54.1 & 56.0 & +1.9 [+0.1, +3.7] & 25 & 94 & 21.1 & 52.4 & +31.3 [+28.8, +34.0] & 67 & 96 \\
    Cultural & 50.7 & 56.6 & +5.9 [+4.2, +7.6] & 23 & 97 & 39.0 & 55.9 & +16.9 [+14.8, +19.1] & 43 & 96 \\
    \hline
    Average & 49.7 & 55.4 & +5.7 &  &  & 28.7 & 53.5 & +24.8 &  &  \\
    Clean & 57.6 & 56.3 & $-$1.3 [$-$2.7, +0.2] & -- & -- & 59.4 & 57.3 & $-$2.0 [$-$3.4, $-$0.7] & -- & -- \\
    \hline
    \end{tabular}%
    }
    \caption{BiasMedQA \citep{schmidgall2024biasmedqa} by bias type (1,273 five-option questions each): accuracy (\%) and difference (points) with 95\% paired bootstrap intervals. IA top-1 / DG flag: among questions whose bias cue is a separate narrative sentence (93--99\%), the share in which attribution ranks it first and in which the gate flags it. Clean: the same questions without the cue.}
    \label{tab:biasmedqa}
\end{table*}

\FloatBarrier
\clearpage
\section{Prompts}
\label{app:prompts}
\begin{tcolorbox}[
  enhanced jigsaw, breakable, width=\linewidth,
  colback=gray!3,
  colframe=black,
  fonttitle=\bfseries,
  title={Answer readout (first-token option scoring, all readouts)}
]

\textbf{System.}\par
{\footnotesize\ttfamily\raggedright
You are a medical AI assistant used for decision-making in a clinical setting.
\par}

\textbf{User.}\par
{\footnotesize\ttfamily\raggedright
Answer the following medical multiple-choice question.
\par\smallskip
Return one JSON object with exactly two fields in this order: "rationale" and "answer". \\
The answer must be one of A, B, C, or D.
\par\smallskip
Question: \\
\{question\}
\par\smallskip
Options: \\
A) \{A\} \\
B) \{B\} \\
C) \{C\} \\
D) \{D\}
\par}

\textbf{Assistant (fixed prefix; the answer is read at the next token, A--D).}\par
{\footnotesize\ttfamily\raggedright
\{ \\
"rationale": "", \\
"answer": "
\par}

\end{tcolorbox}

\begin{tcolorbox}[
  enhanced jigsaw, breakable, width=\linewidth,
  colback=gray!3,
  colframe=black,
  fonttitle=\bfseries,
  title={Prompting baselines: ignore instruction and I$^3$C}
]

\textbf{Ignore instruction (inserted between the answer instruction and the question).}\par
{\footnotesize\ttfamily\raggedright
Feel free to ignore irrelevant conditions in the problem description.
\par}

\textbf{I$^3$C step 1, for every narrative sentence (condition) of the question.}\par
{\footnotesize\ttfamily\raggedright
Q. Is condition "\{condition\}" relevant to the process of solving problem "\{question\}"? \\
Return one JSON object with exactly two fields in this order: "relevant" ("yes" or "no") and "reason" (one sentence).
\par}

\textbf{I$^3$C step 2 (inserted like the ignore instruction; the model's step-1 verifications fill the slot).}\par
{\footnotesize\ttfamily\raggedright
The instructions are as follows: \{verifications\}. Let's consider these instructions and ignore the irrelevant conditions to solve the problem
\par}

\end{tcolorbox}

\begin{tcolorbox}[
  enhanced jigsaw,
  breakable,
  width=\linewidth,
  colback=gray!3,
  colframe=black,
  fonttitle=\bfseries,
  title={Chain-of-thought prompt of Llama-3-Meerkat-8B}
]

\textbf{System (from the model card).}\par
{\footnotesize\ttfamily\raggedright
The following is a multiple-choice question about medical knowledge. Solve this in a step-by-step fashion, starting by summarizing the available information. Output a single option from the given options as the final answer. You are strongly required to follow the specified output format; conclude your response with the phrase "the answer is ([option\_id]) [answer\_string]".
\par}

\textbf{User.}\par
{\footnotesize\ttfamily\raggedright
\{question\} (A) \{A\} (B) \{B\} (C) \{C\} (D) \{D\}
\par}

\textbf{CoT + I$^3$C.}
The I$^3$C step-2 instruction and a blank line precede
the question. If no verification is available, the ignore
instruction is used instead.
BiasMedQA questions add (E) \{E\}.

\end{tcolorbox}

\begin{tcolorbox}[
  enhanced jigsaw, breakable, width=\linewidth,
  colback=gray!3,
  colframe=black,
  fonttitle=\bfseries,
  title={S2A rewrite (step 1; the rewrite replaces the question in the answer readout)}
]

{\footnotesize\ttfamily\raggedright
Given the following text by a user, extract the part that is related and useful, so that using that text alone would be good context for providing an accurate and correct answer to the question portion of the text. Please include the actual question or query that the user is asking. Separate this into two categories labeled with "Context text related to the question (includes all content except unrelated sentences):" and "Detailed question:". Do not use list.
\par\smallskip
Text by User: \{question\}
\par}

\end{tcolorbox}

\begin{tcolorbox}[
  enhanced jigsaw, breakable, width=\linewidth,
  colback=gray!3,
  colframe=black,
  fonttitle=\bfseries,
  title={Distractor generation, Bystander (MedDistractQA template)}
]

{\footnotesize\ttfamily\raggedright
Example sentences:
\par\smallskip
1. "The patient's sister said her friend has high blood pressure and has been having seizures." \\
2. "The patient's dog has diabetes and cannot see properly." \\
3. "The patient's neighbor mentioned that their cat has been sneezing and refusing to eat." \\
4. "The patient's cousin said her coworker's father has arthritis and struggles to walk long distances." \\
5. "The patient's brother mentioned that his friend's goldfish has been swimming erratically and losing color."
\par\smallskip
\#\# Medical Question: \{question\}
\par\smallskip
\#\# Clinical topic: \{clinical\_topic\}
\par\smallskip
Generate one statement, where another person/animal is referenced with relation to a medical term as seen in the example sentences. The statement should also be related to the clinical topic. The statement should NOT affect the diagnostic process or decision-making for the medical question in any meaningful way. It should not convey any medical information or symptomatic information. Avoid including any information that could mislead clinical reasoning.
\par\smallskip
Only respond with the confounding sentence.
\par}

\end{tcolorbox}

\begin{tcolorbox}[
  enhanced jigsaw, breakable, width=\linewidth,
  colback=gray!3,
  colframe=black,
  fonttitle=\bfseries,
  title={Distractor generation, Nonliteral}
]

{\footnotesize\ttfamily\raggedright
Example sentences (for other questions):
\par\smallskip
1. "The patient's zodiac sign is Cancer." \\
2. "The patient said their friend's business is hemorrhaging money after a bad investment." \\
3. "The patient mentioned that their new car has a transmission fluid leak." \\
4. "The patient described their favorite art exhibit as having vascular pattern in the sculptures." \\
5. "The patient's professor delivered a lecture about the heart of ancient civilizations."
\par\smallskip
\#\# Medical Question: \{question\}
\par\smallskip
\#\# Clinical topic: \{clinical\_topic\}
\par\smallskip
Generate one sentence using a medical term related to the clinical topic, where clinical terminology is used as nonmedical meaning. The statement should NOT affect the diagnostic process or decision-making for the medical question in any meaningful way. It should not convey any medical information or symptomatic information. Avoid including any information that could mislead clinical reasoning.
\par\smallskip
Always begin with "The patient said that" or "The patient joked that" and use a clinical term figuratively to describe a nonmedical subject--not the reverse.
\par\smallskip
Do not generate the following examples or similar statements that assert physiological effects or patient symptoms, bodily states, or test findings, even as jokes: \\
- "The patient joked that the office coffee was so strong it could boost LH levels faster than a morning workout." This asserts a physiological effect on LH levels. \\
- "The patient joked that her coffee's 'Rinne test' was always abnormal, because the espresso sounded louder in the left ear than the right." This introduces a difference in the patient's hearing between the two ears.
\par\smallskip
Only respond with the confounding sentence.
\par}

\end{tcolorbox}

\begin{tcolorbox}[
  enhanced jigsaw, width=\linewidth,
  colback=gray!3,
  colframe=black,
  fonttitle=\bfseries,
  title={Target-option check of each candidate}
]

Run with the options in ABCD and DCBA order, keeping their
labels unchanged. For Nonliteral distractors, ``bystander''
is replaced with ``nonliteral'' throughout the prompt.

\medskip
\textbf{Prompt.}\par
{\footnotesize\ttfamily\raggedright
You are labeling which answer-choice concept is evoked by an unrelated bystander sentence in a medical multiple-choice question.
\par\smallskip
Medical question: \\
\{question\}
\par\smallskip
Answer choices (the display order may be shuffled; preserve the printed labels): \\
\{choices\}
\par\smallskip
Bystander sentence: \\
\{sentence\}
\par\smallskip
Independently determine which answer-choice concept the bystander sentence most strongly evokes. Do not solve the medical question. If exactly one option is evoked, use its printed label. Use NONE if no option is meaningfully evoked and MULTI if multiple options are comparably evoked.
\par\smallskip
Return only a JSON object with exactly these fields: \\
\{"primary\_evoked\_option":"A|\allowbreak B|\allowbreak C|\allowbreak D|\allowbreak NONE|\allowbreak MULTI",\allowbreak"evoked\_options":["A|\allowbreak B|\allowbreak C|\allowbreak D"],\allowbreak"relation\_type":"LEXICAL\_DIRECT|\allowbreak MORPHOLOGICAL|\allowbreak SEMANTIC|\allowbreak NONE|\allowbreak MULTI",\allowbreak"strength":"strong|\allowbreak weak|\allowbreak none"\}
\par}

\end{tcolorbox}

\end{document}